\documentclass[journal]{IEEEtran} 

\usepackage{mathtools}
\usepackage{amsmath,amsfonts}
\usepackage{pifont}

\usepackage{array}
\usepackage[caption=false,font=footnotesize,labelfont=rm,textfont=rm]{subfig}

\usepackage{textcomp}
\usepackage{stfloats}
\usepackage{url}
\usepackage{verbatim}
\usepackage{graphicx}
\usepackage{cite}
\usepackage{booktabs}
\usepackage{multirow}
\usepackage[normalem]{ulem}
\usepackage{float}
\usepackage{threeparttable}
\usepackage{amssymb}

\usepackage[colorlinks,
linkcolor=blue,
anchorcolor=blue,
citecolor=blue]{hyperref}

\DeclareMathAlphabet\mathbfcal{OMS}{cmsy}{b}{n}

\usepackage{soul}

\begin{document}

\title{Safety-Critical LiDAR-Inertial Odometry With On-Manifold Deterministic Protection Level}

\author{Yueqi Zhu, Yan Pan, Chufan Rui, Jiasheng Luo and Bo Zhou
	\thanks{This work was supported by the National Natural Science Foundation (NNSF) of China under Grant 62573124. (Corresponding author: Bo Zhou.)}\thanks{The authors are with the School of Automation, Southeast University and the Key Laboratory of Measurement and Control of CSE, Ministry of Education, Nanjing 210096, P. R. China (email: zhuyq@seu.edu.cn; yanpan@seu.edu.cn; rcf@seu.edu.cn; jasonluo@seu.edu.cn; zhoubo@seu.edu.cn)}
}

\maketitle

\begin{abstract}
In safety-critical scenarios, the protection level of the autonomous navigation system is crucial for enabling mobile robots to perform safe tasks. However, existing studies on probabilistic navigation systems for robots usually perform offline accuracy evaluations using limited datasets and assume that the results can be applied to unknown real-world environments. As a result, current autonomous mobile robots often lack protection levels for online safety assessment. To fill this gap, we propose a safety-critical LiDAR-inertial odometry (LIO) that provides deterministic protection levels based on a novel on-manifold set-membership filter. By adopting the unknown but bounded assumption, we derive a neat closed-form relationship between point cloud noise and the uncertainty of the estimation from the iterative closest point algorithm. Using this relationship, we design an on-manifold ellipsoidal set-membership filter (OM-ESMF) and implement it within the LIO system. Leveraging the properties of the set-membership filter, our system offers the feasible sets of the estimated locations as the deterministic protection levels, serving as safety references for the robots' downstream autonomous operations. The experimental results show that our system can provide effective deterministic online safety references for diverse robots in various environments.
\end{abstract}


\section{Introduction}
Although a variety of mobile robots have been widely deployed to perform various tasks, ensuring the safety of robots, individuals, environments, and interactive objects in challenging environments remains an open problem. In safety-critical scenarios \cite{pappas_2024}, the navigation system plays a crucial role in guaranteeing the safe operation of autonomous mobile robots. Although safety is a decisive factor in the application of autonomous mobile robots, implementing the online safety and reliability assessment of the navigation system is an overlooked challenge. In the field of the global navigation satellite system (GNSS), the protection level (PL) \cite{berbineau_2018} is leveraged as a key metric to quantify the online uncertainty of the navigation system, characterizing the reliability of the locations and providing a safety assessment for downstream autonomous behaviors. However, for widely used navigation systems for mobile robots \cite{fast_lio2, lio_sam}, most of the research focused on improving the accuracy of localization and mapping, while studies on the uncertainty and the protection level are commonly overlooked.

Probabilistic state estimation is a commonly used method for enabling autonomous navigation of mobile robots. The navigation system based on Bayesian filters \cite{fast_lio2, lio_ekf} simultaneously estimates the means and covariances of the system state. The reliability of the navigation system can be measured via the probabilistic protection levels, which are the scaled covariances using the chi-square distribution. However, the estimated protection levels are usually rather overconfident and cannot accurately represent the error between the localization result and the ground-truth location. This limitation arises because the noise distributions of the measurements from the sensors are usually complex and unknown, and probabilistic methods usually assume them to be zero-mean Gaussian distributions with extremely small covariances. The accuracy of this method in modeling noise is crude, which in turn leads to low accuracy in the resulting safety assessment. What's more, common probabilistic methods are prone to overlooking the uncertainties of submaps \cite{tang_2023}. They trust the observation between the current frame and the current submap without addressing the uncertainty and drift of the submap. Optimization-based navigation systems \cite{orb_slam3, lio_sam} take the optimal result as an estimation and generally do not account for the uncertainty and reliability of the estimated locations. During the design phase of the optimization-based navigation systems, offline performance evaluation for accuracy is commonly conducted using datasets with reliable ground truth, whereas online performance evaluation under various real-world environments without ground truth is rarely considered. Nevertheless, although various datasets for challenging scenarios were proposed \cite{subt-mrs, mars-lvig}, the number and complexity of the scenarios in these datasets can never cover all the situations encountered in the real world. As a result, offline accuracy evaluation results based on datasets cannot truly reflect the performance of navigation systems under all actual working conditions. This offline absolute trust may pose potential safety hazards. 

Unlike probabilistic state estimation, deterministic state estimation \cite{calafiore2005, rohou_2023} is a more robust type, which can estimate the online bounded uncertainty of the system state without involving probability. The set-membership filter (SMF) \cite{wang_2022, yang_2023} is a typical deterministic state estimation tool. It only requires that the noise satisfy the unknown but bounded (UBB) assumption, which can simply and accurately describe complex sensor noises that are difficult to calibrate. Moreover, a set-membership filter estimates the feasible set of the system state and directly provides the uncertainty and reliability of the estimated state. However, due to the practical complexity of the set arithmetic and the theoretical complexity of robot systems on high-dimensional manifolds, very few studies have extended the set-membership filter to the high-dimensional robot autonomous navigation system using 3D LiDAR and 6-axis inertial measurement unit (IMU).

To overcome the aforementioned robot safety issue, we propose to use the feasible set from the set-membership filter as the deterministic protection level. Compared with the protection level based on the unbounded Gaussian distribution, the deterministic protection level characterizes the uncertainties of the localization results in bounded spaces. To break through the theoretical limitations of the deterministic state estimation in safety-critical scenarios and explore its application potential, in this paper, we present a LiDAR-inertial odometry (LIO) based on a meticulously designed on-manifold ellipsoidal set-membership filter. By designing an on-manifold set-membership filter, our system can provide protection levels in the form of ellipsoidal sets and offer more accurate online safety assessments. To the best of our knowledge, this is the first open-source work to extend the deterministic state estimation to the high-dimensional on-manifold robot system with the purpose of ensuring the safety of autonomous navigation. Our main contributions are summarized as follows:
\begin{enumerate}
	\item{A \emph{set-membership-based deterministic protection level} representation is proposed for safety-critical localization and mapping. The proposed deterministic protection level effectively characterizes the relationship between estimated locations and the ground truth in bounded spaces and provides bounds on the unknown ground-truth locations.}
	
	\item{A distinctive and neat closed-form expression for describing the online reliability of point-to-plane iterative closest point (ICP) results is derived based on the UBB assumption, Lie groups, and Lie algebras.}
	
	\item{A novel \emph{on-manifold ellipsoidal set-membership filter (OM-ESMF)} is designed to fuse the observations from the IMU with the ICP results. The final output of this filter is the ellipsoidal set where the system state lies. The resultant ellipsoidal sets can serve as effective and online deterministic protection levels.}
	
	\item{Experiments conducted with multiple mobile robot platforms demonstrate that the estimated protection level from our system outperforms the state-of-the-art (SOTA) methods in terms of accuracy and effectiveness. In our experiments, the cover rate of the protection levels estimated by our system to the ground truth reached 100\%.}
\end{enumerate}

Our code is publicly available \footnote{[Online]. Available: https://github.com/Zhu-YQ/LIO-SMF.git.}.

\section{Related Works}
\subsection{State Estimation for Robotics}
As the well-established route, the probabilistic state estimation \cite{barfoot_2017} has been widely applied to robots \cite{fast_lio2, orb_slam3}. Unlike probabilistic state estimation, deterministic state estimation is another promising approach. As the pivotal method of deterministic state estimation, the set-membership filter \cite{qin_2024, zhu_2024} uses bounded sets to describe the uncertainties within the systems. In terms of the selection of sets, the ellipsoidal set \cite{schweppe_1968, walter_2001} is a widely used choice. To improve the performance of ellipsoidal SMF, Wang \textit{et al.}~\cite{wang_2022} proposed an optimal ellipsoidal SMF with less conservatism by assuming that the inverse of the observation model exists. Wang \textit{et al.}~\cite{wang_2023} increased the degrees of freedom in the SMF, enabling the filter to better adapt to the dynamic changes of the system. This method not only improved the estimation accuracy but also reduced the computational load.

Although there are numerous theoretical studies related to the SMF, the corresponding works focusing on real-world applications are relatively rare. Zhu \textit{et al.}~\cite{zhu_2023} applied the ellipsoidal SMF to the two-dimensional mobile robot localization based on ultrasonic sensors. Zhou \textit{et al.}~\cite{11081890} proposed a method for applying the ellipsoidal SMF to three-dimensional ultra-wideband localization. Moreover, although SMFs have unique characteristics in the field of state estimation and possess high potential application value, very few studies have applied SMFs to state estimation problems based on manifolds \cite{fast_lio2}. To enable the SMF to demonstrate its advantages in on-manifold state estimation, in this paper, we design an on-manifold SMF and combine it with the uncertainty modeling in the LIO system, presenting a feasible SMF-based LIO.

\subsection{Uncertainty Analysis for LiDAR-Based Localization}
The safety threats of the localization system are usually caused by the uncertainties within the system. Therefore, uncertainty analysis \cite{arevalo_2018} is of vital importance in constructing a safety-critical localization system. Censi~\cite{censi_2007} systematically analyzed the error sources of the ICP algorithm. Based on the implicit function theorem, this work provided a covariance estimation for the ICP algorithm results. However, this method is only applicable to the two-dimensional ICP algorithm. Prakhya \textit{et al.}~\cite{lin_2015} extended this method to the three-dimensional ICP problem and provided a more intuitive closed-form solution. However, this result was derived based on Euler angles and is not applicable to modern state estimation systems constructed based on SE(3). Brossard \textit{et al.}~\cite{brossard_2020} argued that ICP uncertainty is intrinsically governed by the initialization accuracy, and employed the unscented transform to jointly model sensor noise, biases, and initialization error along with their correlations. Unfortunately, the aforementioned methods pay less attention to modeling the point cloud noise, which consequently leads to lower accuracy of the covariance estimations.

Based on the characteristics of LiDAR, Yuan \textit{et al.}~\cite{yuan_2021} constructed an on-manifold probabilistic noise model for measured points, which has become a widely used modeling method for point cloud noise \cite{liu_2024, dong_2025}. Furthermore, Yuan \textit{et al.}~\cite{zhang_2022} explicitly presented a relationship between point uncertainty and the uncertainty of the estimated plane for ICP, thereby presenting the uncertainty of the map. Huang \textit{et al.}~\cite{huang_2024} achieved a more accurate description of the probabilistic point noise model by additionally considering the incident angle and the surface roughness. Furthermore, they also proposed an incremental method for maintaining map uncertainty, which significantly improved the efficiency of updating map uncertainty. Hu \textit{et al.}~\cite{paloc} presented an optimization-based LiDAR localization system with uncertainty estimation. By combining factor graphs and the Laplace approximation \cite{jimenez_2019}, this method can provide considerable online reliability assessment.

\section{Preliminaries on Ellipsoidal Sets}
\label{sec:preliminary}
Since the ellipsoidal sets are chosen to describe the UBB noises and uncertainties, there are numerous ellipsoidal set operations in our system. In this section, we list the necessary and mature operations of ellipsoidal sets in order to introduce our system in the following sections better.

\subsection{Definition}
An $n$-dimensional ellipsoidal set is defined as follows:
\begin{equation}
	\mathcal{E} \left( \mathbf{a},\mathbf{P} \right) =\left\{ \mathbf{x}\mid \left( \mathbf{x}-\mathbf{a} \right) ^{\mathrm{T}}\mathbf{P}^{-1}\left( \mathbf{x}-\mathbf{a} \right) \leqslant 1, \ \mathbf{x} \in \mathbb{R}^n \right\} 
\end{equation}
where the vector $\mathbf{a} \in \mathbb{R}^n$ is called the center, and the positive-definite matrix $\mathbf{P} \in \mathbb{R} ^{n\times n}$ is called the shape matrix.

\subsection{Linear Map and Translation}
\label{sec:linea_map}
Linear map and translation are the basic operations of vectors and ellipsoidal sets. If the original $n$-dimensional vector satisfies $\mathbf{x}\in \mathcal{E} \left( \mathbf{a},\mathbf{P} \right) $, the transformed vector satisfies:
\begin{equation}
	\mathbf{Ax}+\mathbf{b}\in \mathcal{E} \left( \mathbf{Aa}+\mathbf{b},\mathbf{APA}^{\mathrm{T}} \right) 
\end{equation}
where $\mathbf{A}\in \mathbb{R} ^{m\times n}$, $\mathbf{b}\in \mathbb{R} ^m$.

\subsection{Minkowski Sum}
\label{sec:minkoeski}
Considering multiple vectors within different ellipsoidal sets $\mathbf{x}_i \in \mathcal{E} \left( \mathbf{a}_i,\mathbf{P}_i \right) $, $i=1,2,...,n$, if the sum of all $\mathbf{x}_i$ is required, the result is within the following Minkowski sum \cite{wang_2022}:
\begin{equation}
	\sum_{i=1}^n{\mathbf{x}_i}\in \bigoplus_{i=1}^n{\mathcal{E} \left( \mathbf{a}_i,\mathbf{P}_i \right)}
\end{equation}
where $\oplus$ is the Minkowski sum operator. However, the resultant Minkowski sum will not necessarily be an ellipsoidal set. For the consistency of the subsequent operations using ellipsoidal sets, the out-bounding ellipsoidal set must be maintained:
\begin{equation}
	\mathcal{E} \left( \mathbf{a},\mathbf{P} \right) \supseteq \bigoplus_{i=1}^n{\mathcal{E} \left( \mathbf{a}_i,\mathbf{P}_i \right)}
\end{equation}
whose parameters are $\mathbf{a}=\sum_{i=1}^n{\mathbf{a}_i},\ \mathbf{P}=\sum_{i=1}^n{\beta _{i}^{-1}\mathbf{P}_i}$ \cite{walter_2001}, where $\beta _i>0$ and $\sum_{i=1}^n{\beta _i}=1$. Moreover, due to the flexibility of $\beta _i$, the out-bounding ellipsoidal set is not unique. To limit the conservatism of the set, the minimum trace criterion \cite{wang_2022} is frequently used to obtain a unique set,
whose closed-form solution is $\beta _i={{\sqrt{\mathrm{tr}\left( \mathbf{P}_i \right)}}/{\sum\nolimits_{j=1}^n{\sqrt{\mathrm{tr}\left( \mathbf{P}_j \right)}}}}$ \cite{walter_2001}. Denote the corresponding optimal ellipsoidal set as $\mathcal{E} \left( \mathbf{a}^*,\mathbf{P}^* \right)$. For clarity, in this paper, the operator $\oplus _{\mathcal{E}}$ is used to represent the operation of obtaining the optimal out-bounding ellipsoidal set of the Minkowski sum, that is
\begin{equation}
	\mathcal{E} \left( \mathbf{a}^*,\mathbf{P}^* \right) =\bigoplus_{i=1}^n{_{\mathcal{E}}\mathcal{E} \left( \mathbf{a}_i,\mathbf{P}_i \right)}
\end{equation}
%\begin{equation}
%	\begin{aligned}
	%		&\mathcal{E} \left( \mathbf{a}^*,\mathbf{P}^* \right) =\bigoplus_{i=1}^n{_{\mathcal{E}}\mathcal{E} \left( \mathbf{a}_i,\mathbf{P}_i \right)}
	%		\\
	%		&=\mathcal{E} \left( \mathbf{a}_1,\mathbf{P}_1 \right) \oplus _{\mathcal{E}}\mathcal{E} \left( \mathbf{a}_2,\mathbf{P}_2 \right) \oplus _{\mathcal{E}}\cdots \oplus _{\mathcal{E}}\mathcal{E} \left( \mathbf{a}_{\mathrm{n}},\mathbf{P}_{\mathrm{n}} \right) 
	%	\end{aligned}
%\end{equation}

\subsection{Intersection}
\label{sec:intersection}
Intersection is a momentous operation for set-membership filters. For two ellipsoidal sets $\mathcal{E} \left( \mathbf{a}_1,\mathbf{P}_1 \right) $ and $\mathcal{E} \left( \mathbf{a}_2,\mathbf{P}_2 \right) $ with the same dimension, their intersection is usually not retained as an ellipsoidal set. With the same handling in the Minkowski sum operation, the out-bounding ellipsoidal set \cite{wang_2022} is required:
\begin{equation}
	\mathcal{E} \left( \mathbf{a},\mathbf{P} \right) \supseteq \left( \mathcal{E} \left( \mathbf{a}_1,\mathbf{P}_1 \right) \cap \mathcal{E} \left( \mathbf{a}_2,\mathbf{P}_2 \right) \right) 
\end{equation}
\begin{equation}
	\mathbf{a}=\mathbf{P}_{\lambda}^{}\left( \left( 1-\lambda \right) \mathbf{P}_{1}^{-1}\mathbf{a}_1+\lambda \mathbf{P}_{2}^{-1}\mathbf{a}_2 \right) ,\ 0<\lambda<1
\end{equation}
\begin{equation}
	\mathbf{P}=\left( 1-\mathrm{\nu} \right) \mathbf{P}_{\lambda}^{}
\end{equation}
where
\begin{equation}
	\mathbf{P}_{\lambda}^{-1}=\left( 1-\lambda \right) \mathbf{P}_{1}^{-1}+\lambda \mathbf{P}_{2}^{-1}
\end{equation}
\begin{equation}
	\nu =\left( 1-\lambda \right) \mathbf{a}_{1}^{\mathrm{T}}\mathbf{P}_{1}^{-1}\mathbf{a}_1+\lambda \mathbf{a}_{2}^{\mathrm{T}}\mathbf{P}_{2}^{-1}\mathbf{a}_2-\mathbf{a}_{}^{\mathrm{T}}\mathbf{P}_{\lambda}^{-1}\mathbf{a}_{}
\end{equation}
Additionally, the minimum trace criterion is also used to obtain a unique set,
%\begin{equation}
%	\underset{\lambda }{\min}\,\,\mathrm{tr}\left( \mathbf{P} \right) 
%\end{equation}
whose solution can be acquired using the linear search methods \cite{wright2018}. Similarly, to enhance the clarity, in this paper, the operator $\cap _{\mathcal{E}}$ is used to represent the operation of obtaining the optimal out-bounding ellipsoidal set with minimum trace of the intersection, that is
\begin{equation}
	\mathcal{E} \left( \mathbf{a}^*,\mathbf{P}^* \right) =\mathcal{E} \left( \mathbf{a}_1,\mathbf{P}_1 \right) \cap _{\mathcal{E}}\mathcal{E} \left( \mathbf{a}_2,\mathbf{P}_2 \right) 
\end{equation}

\subsection{Conversion Between Box and Ellipsoidal Set}
%\subsection{Conversion from box to ellipsoidal set}
\label{sec:box}
A box \cite{Merlinge_2024} is composed of multiple intervals, and this representation for UBB vectors is more intuitive and suitable to be used as an auxiliary set for performing ellipsoidal set operations.

An $n$-dimensional box is defined as follows:
\begin{equation}
	\mathcal{B} =\left[ b_1 \right] \times \left[ b_2 \right] \times \cdots \times \left[ b_{\mathrm{n}} \right] 
\end{equation}
\begin{equation}
	\left[ b_i \right] =\left\{ x\mid \mathrm{inf}\left( \left[ b_i \right] \right) \leqslant x\leqslant \mathrm{sup}\left( \left[ b_i \right] \right) ,x\in \mathbb{R} \right\}
\end{equation}
where $i=1,2,...,n$, $\times$ is the Cartesian product operator, $[b_i]$ is an interval \cite{ehambram_2022}. 

Considering an $n$-dimensional ellipsoidal set $\mathcal{E} \left( \mathbf{a},\mathbf{P} \right) $, the relationship between it and its inscribed box $\mathcal{B} _{in}$ is
\begin{equation}
	\mathbf{a}=\left[ \begin{matrix}
		a_1&		\cdots&		a_n\\
	\end{matrix} \right] ^{\mathrm{T}}, \ \mathbf{P}=\mathrm{diag}\left( nr_{1}^{2},\cdots ,nr_{n}^{2} \right) 
\end{equation}
\begin{equation}
	a_i=0.5\left( \mathrm{inf}\left( \left[ b_i^\mathrm{in} \right] \right) +\mathrm{sup}\left( \left[ b_i^\mathrm{in} \right] \right) \right),\  r_i=\mathrm{sup}\left( \left[ b_i^\mathrm{in} \right] \right) -a_i
\end{equation}
where
\begin{equation}
	\mathcal{B} _\mathrm{in}=\left[ b_{1}^\mathrm{in} \right] \times \left[ b_{2}^\mathrm{in} \right] \times \cdots \times \left[ b_{n}^\mathrm{in} \right] 
\end{equation}

Conversely, the out-bounding box $\mathcal{B}_{out}$ of the known $\mathcal{E} \left( \mathbf{a}',\mathbf{P}' \right) $ is determined as follows:
\begin{equation}
	\mathcal{B}_\mathrm{out} =\left[ b_1^\mathrm{out} \right] \times \left[ b_2^\mathrm{out} \right] \times \cdots \times \left[ b_{\mathrm{n}}^\mathrm{out} \right] 
\end{equation}
\begin{equation}
	\mathrm{inf}\left( \left[ b_i^\mathrm{out} \right] \right) ={a_i}'-\sqrt{\mathbf{P}'\left( i,i \right)}
\end{equation}
\begin{equation}
	\mathrm{sup}\left( \left[ b_i^\mathrm{out} \right] \right) ={a_i}'+\sqrt{\mathbf{P}'\left( i,i \right)}
\end{equation}
where ${a_i}'$ is the $i$-th element of $\mathbf{a}'$, and $\mathbf{P}'\left( i,i \right)$ indicates the element on the $i$-th row and $i$-th column.

\section{System Overview}
\begin{figure*}[htp]
	\centering
	\includegraphics[width=6in]{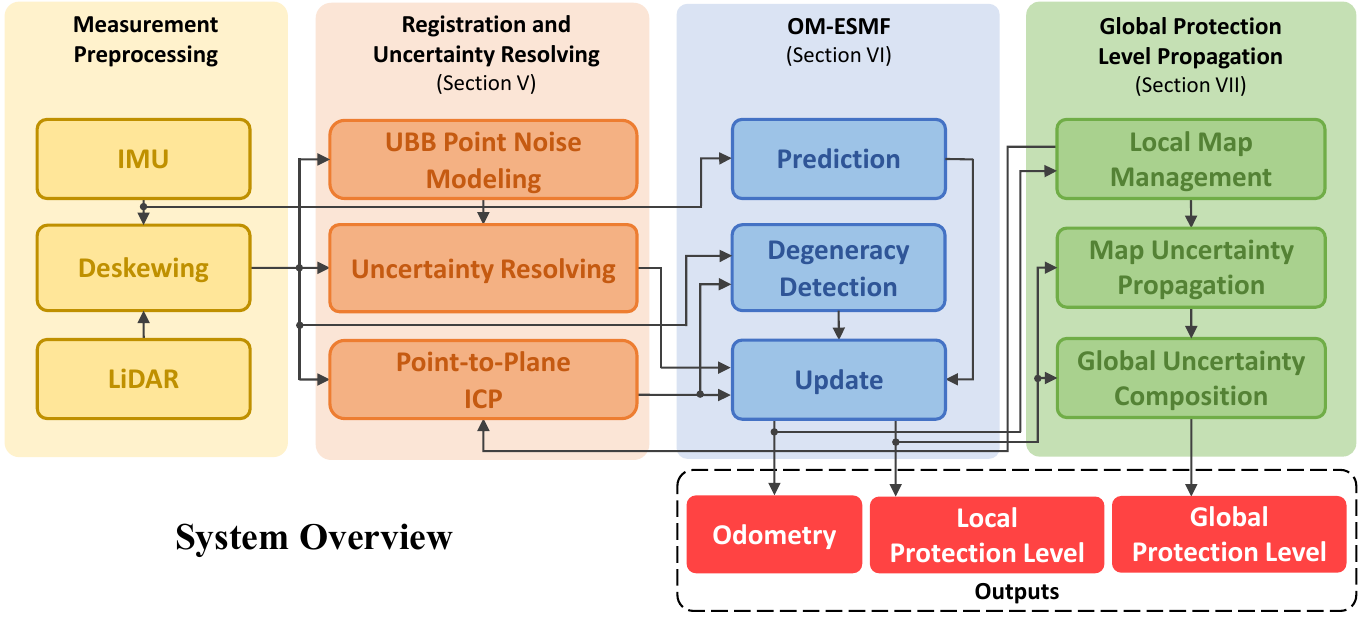}
	\caption{Overview of the proposed safety-critical LiDAR-inertial odometry with deterministic protection levels. The IMU measurements support scan deskewing and the \emph{OM-ESMF} prediction, while the deskewed LiDAR points are modeled with UBB range and bearing errors and registered to the active local map by point-to-plane ICP. The uncertainty-resolving method propagates the bounded point errors into a pose-error ellipsoidal set in the same product-manifold coordinates used by the filter. The \emph{OM-ESMF} fuses this ICP observation set with the IMU-predicted sets and applies a constrained update in ill-conditioned directions, yielding the odometry and \emph{local protection levels}. Reliable scans update the active local map. At a map switch, the posterior pose set is stored as the relative map-anchor uncertainty, which is recursively propagated and composed with the current local pose set to obtain the \emph{global protection levels}.}
	\label{fig:pipeline}
\end{figure*}

The overall pipeline is illustrated in Fig. \ref{fig:pipeline}. It consists of four coupled modules: \emph{measurement preprocessing}, \emph{registration and uncertainty resolving}, \emph{on-manifold ellipsoidal set-membership filter}, and \emph{global protection level propagation}.

The \emph{measurement preprocessing} module receives the raw LiDAR scans and IMU measurements. The IMU measurements are first used to deskew each scan and are also supplied to the prediction stage of the \emph{OM-ESMF}.

In the \emph{registration and uncertainty resolving} module, the deskewed point cloud is used by the scan-to-map point-to-plane ICP, which aligns the current scan with the active local map. For every deskewed LiDAR point, the range and bearing errors are modeled as UBB quantities and converted into a three-dimensional point-error ellipsoidal set. The proposed uncertainty-resolving method propagates the bounded point errors through the locally converged ICP problem and produces an ellipsoidal set of the additive translation and right rotation errors. Hence, the registration module provides the filter with a pose observation and its associated deterministic uncertainty set rather than a pose alone.

The \emph{OM-ESMF} fuses the two information sources. The IMU branch propagates the nominal translation, velocity, and rotation together with their error sets, while the LiDAR branch directly supplies observation sets in the same error coordinates. Degeneracy is detected from the localizability of the point-to-plane ICP problem before the update. When ill-conditioned directions are detected, the box-based constrained update retains the predicted bounds along those directions and only uses the LiDAR observation in sufficiently constrained directions. After state injection and error-set recentering, the \emph{OM-ESMF} outputs the odometry pose and the \emph{local protection levels} with respect to the active local-map frame.

The \emph{global protection level propagation} module accounts for the fact that the active local map is itself anchored by an uncertain historical pose. During a local-map period, reliable registered scans are transformed with the updated pose and inserted into the active map. When the map-switching condition is satisfied, the posterior pose sets at the switching instant are stored as the uncertainties of the relative translation and rotation between the old and new local-map frames. These translational and rotational sets are propagated separately through successive pose compositions while retaining their coupling. At each time step, the propagated map-anchor sets are composed with the current local pose sets. The resulting global translation and rotation sets are the \emph{global protection levels}.

\section{Point Cloud Registration With UBB Uncertainty Resolving}
\label{sec:uncertainty_resolving}

\subsection{Point Model With UBB Noise}
In this section, we assume that the collected point cloud is deskewed using the IMU measurements \cite{lio_sam}. Denote the reference frame of the IMU as $\{I\}$, and the reference frame of the LiDAR as $\{L\}$. The relationship between the measured point from the laser and the ground-truth point can be formulated as
\begin{equation}
	{^L\mathbf{p}_i}={^L\tilde{\mathbf{p}}_i}+{^L\delta \mathbf{p}_i}
	\label{eq:point_R3}
\end{equation}
where ${^L\mathbf{p}_i} \in \mathbb{R}^3$ is the ground-truth point, ${^L\tilde{\mathbf{p}}_i} \in \mathbb{R}^3$ is the measured point, and ${^L\delta \mathbf{p}_i} \in \mathbb{R}^3$ is the unknown noise. Since ${^L\delta \mathbf{p}_i}$ is unknown, it is commonly modeled as a random variable following a zero-mean Gaussian distribution. However, the Gaussian distribution is unbounded, whereas the noise captured by real-world sensors is bounded. This suggests that Gaussian distributions are inevitably an approximation for modeling noise. In addition, the noise characteristics of the point vary at different ranges. To accurately characterize the noise of a point using the Gaussian distribution, it is necessary to obtain different covariances corresponding to different ranges. However, it is intractable and impractical to accurately calibrate the dynamic covariance of the noise at different ranges due to the complexity of the hardware and environment. Existing methods \cite{fast_lio2, lio_ekf} usually set the covariance of the noise to a small empirical value. Thus, the accuracy of existing methods for modeling noise based on Gaussian distributions is limited. 

%\begin{figure}[tp]
%	\centering
%	\includegraphics[width=3in]{pics/point}
%	\caption{An illustration of the point model with UBB noise. The unknown ground-truth point is encompassed by a bounded ellipsoidal set whose center is the measured point.}
%	\label{fig:point}
%\end{figure}

\begin{figure}[tp]
	\centering
	\includegraphics[width=3.5in]{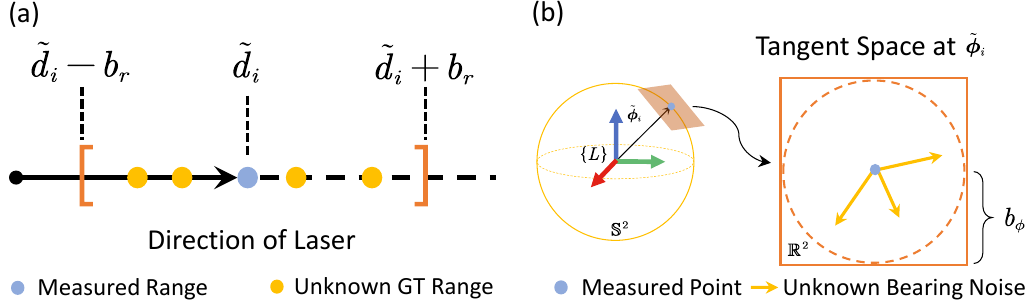}
	\caption{An illustration of the point model with UBB noises. (a) Range measurement with UBB noise. The unknown ground-truth range is restricted within a bounded set. (b) Bearing measurement with UBB noise. The length of the unknown bearing noise is less than the upper bound.}
	\label{fig:range_and_bearing}
\end{figure}

To overcome this drawback, we utilize the UBB assumption and the ellipsoidal set to model the noise. Originally, based on the theory in \cite{yuan_2021}, the measurement of a laser is composed of a range and a bearing:
\begin{equation}
	d_i=\tilde{d}_i+n_{i}^{\left( d \right)} ,\ \boldsymbol{\phi }_i=\tilde{\boldsymbol{\phi}}_i\boxplus _{\mathbb{S} ^2}\boldsymbol{n}_{i}^{\left( \phi \right)}
	\label{eq:point_sep}
\end{equation}
where $d_i \in \mathbb{R}$ and $\boldsymbol{\phi }_i \in {\mathbb{S} ^2}$ are the ground-truth range and bearing, $\tilde{d}_i \in \mathbb{R}$ and $\tilde{\boldsymbol{\phi}}_i \in {\mathbb{S} ^2}$ are the measured range and bearing, $n_{i}^{\left( d \right)} \in \mathbb{R}$ and $\boldsymbol{n}_{i}^{\left( \phi \right)} \in \mathbb{R}^2$ are the UBB noises, and $\boxplus _{\mathbb{S} ^2}$ indicates the operation of adding a perturbation. As depicted in Fig. \ref{fig:range_and_bearing}, in our system, $n_{i}^{\left( d \right)}$ and $\boldsymbol{n}_{i}^{\left( \phi \right)}$ are assumed to be UBB, that is $n_{i}^{\left( d \right)}\in \left[ -b_r,b_r \right] $ and $\| \boldsymbol{n}_{i}^{\left( \phi \right)} \| \leqslant b_{\phi}$. The value of $n_{i}^{\left( d \right)}$ indicates the difference between the measured range and the ground-truth range, and the length of $\boldsymbol{n}_{i}^{\left( \phi \right)}$ indicates the angle between the measured bearing and the ground-truth bearing. Unlike the noise modeling methods with Gaussian distributions, when using the UBB assumption for noise modeling, there is no need to provide the covariance. Instead, only the upper bounds of the range error and the bearing error need to be provided, which is a more relaxed and simpler assumption. Consequently, according to the rule of out-bounding ellipsoidal set (see Section \ref{sec:box}), the augmented noise is restricted in the following set:
\begin{equation}
	\left[ \begin{matrix}
			n_{i}^{\left( d \right)}	\ \	{\boldsymbol{n}_{i}^{\left( \phi \right)}}^{\mathrm{T}}\\
		\end{matrix} \right] ^{\mathrm{T}} \in  \mathcal{E} \left( \mathbf{0},^P\mathbf{P}_{i}^{\mathrm{p}} \right) \triangleq \mathcal{E} \left( \mathbf{0},\mathrm{diag}\left( 3b_{r}^{2},3b_{\phi}^{2},3b_{\phi}^{2} \right) \right) 
\end{equation}
By composing the range component and the bearing component, based on the method in \cite{yuan_2021}, \eqref{eq:point_R3} can be reformulated to
\begin{equation}
	\begin{aligned}
		^L\mathbf{p}_i&=d_i\boldsymbol{\phi }_i
		\\
		&=\underset{^L\tilde{\mathbf{p}}_i}{\underbrace{\tilde{d}_i\tilde{\boldsymbol{\phi}}_i}}+\underset{^L\delta \mathbf{p}_i}{\underbrace{\left[ \begin{matrix}
					\tilde{\boldsymbol{\phi}}_i&		-\\
				\end{matrix}\tilde{d}_i\tilde{\boldsymbol{\phi}}_{i}^{\land}\mathbf{N}\left( \tilde{\boldsymbol{\phi}}_i \right) \right] \left[ \begin{array}{c}
					n_{i}^{\left( d \right)}\\
					\boldsymbol{n}_{i}^{\left( \phi \right)}\\
				\end{array} \right] +\mathbf{r}_{\mathrm{p}}^{\mathrm{nl}}}}
	\end{aligned}
\end{equation}
where $( \cdot) ^{\land}$ denotes the skew-symmetric operator, $\mathbf{N}( \tilde{\boldsymbol{\phi}}_i ) \in \mathbb{R} ^{3\times 2}$ is the orthonormal basis of the tangent plane at $\tilde{\boldsymbol{\phi}}_i$, which can be obtained through the Gram-Schmidt orthogonalization procedure, $\mathbf{r}_{\mathrm{p}}^{\mathrm{nl}}\in \mathcal{E} \left( \mathbf{0},\mathbf{P}_{\mathrm{p}}^{\mathrm{nl}} \right) $ is the UBB remainder \cite{scholte_2003}. The derivation is given in Appendix~\ref{sec:appd1}. Denote that $\mathbf{A}_i=[ \begin{matrix}
	\tilde{\boldsymbol{\phi}}_i&		-\\
\end{matrix}\tilde{d}_i\tilde{\boldsymbol{\phi}}_{i}^{\land}\mathbf{N}( \tilde{\boldsymbol{\phi}}_i )] $, then the noise can be given by the rule of linear map (see Section \ref{sec:linea_map}):
\begin{equation}
	^L\delta \mathbf{p}_i\in \mathcal{E} \left( \mathbf{0},{^{L}\mathbf{P}}_{i}^{\mathrm{p}} \right) \triangleq \mathcal{E} \left( \mathbf{0},{\mathbf{A}_i}^P\mathbf{P}_{i}^{\mathrm{p}}\mathbf{A}_{i}^{\mathrm{T}} \right) \oplus _{\mathcal{E}}\mathcal{E} \left( \mathbf{0},\mathbf{P}_{\mathrm{p}}^{\mathrm{nl}} \right) 
\end{equation}
where $\mathbf{P}_{\mathrm{p}}^{\mathrm{nl}}$ is calibrated offline using the Monte Carlo sampling \cite{cong_2022} over the prescribed UBB ranges of the range and bearing errors and the LiDAR operating domain. For each sample, the exact nonlinear remainder is evaluated, and the largest sampled norm is enlarged by a safety factor greater than one to construct an empirical isotropic remainder ellipsoidal set. This conservative method is used for all the following remainders.

Finally, the overall point model written in terms of $\{I\}$ can be formulated as
\begin{equation}
	\begin{aligned}
		{^I\mathbf{p}}_i={_{L}^{I}\mathbf{R}}{^L\mathbf{p}}_i+{_{L}^{I}\mathbf{t}}&={_{L}^{I}\mathbf{R}}\left( {^L\tilde{\mathbf{p}}_i}+{^L\delta \mathbf{p}_i} \right) +{_{L}^{I}\mathbf{t}}
		\\
		&=\underset{^I\tilde{\mathbf{p}}_i}{\underbrace{\left( _{L}^{I}\mathbf{R}^L\tilde{\mathbf{p}}_i+{_{L}^{I}\mathbf{t}} \right) }}+{_{L}^{I}\mathbf{R}}{^L\delta \mathbf{p}_i}
	\end{aligned}	
\end{equation}
\begin{equation}
	{^I\mathbf{p}_i}-{^I\tilde{\mathbf{p}}_i}={_{L}^{I}\mathbf{R}}{^L\delta \mathbf{p}_i}\in \mathcal{E} (\mathbf{0},{^I\mathbf{P}}_{i}^{\mathrm{p}})\triangleq \mathcal{E} (\mathbf{0},_{L}^{I}\mathbf{R}^L{\mathbf{P}_{i}^{\mathrm{p}}}{_{L}^{I}\mathbf{R}}^{\mathrm{T}})
	\label{eq:final_point}
\end{equation}
where ${_{L}^{I}\mathbf{R}} \in \mathrm{SO}(3)$ and ${^{I}_{L}\mathbf{t}}\in \mathbb{R} ^3$ are the calibrated extrinsic parameters. Due to the maturity of the calibration of the extrinsic parameters, the accuracy impact brought by them on the system is small enough to be neglected. For the simplicity of our derivation, the uncertainties of the extrinsic parameters are ignored. This point model with UBB noise is further utilized in Section \ref{sec:uncertainty_resolving_B} to derive the uncertainty of the estimated pose from the ICP algorithm.

\subsection{Closed-Form Uncertainty Resolving}
\label{sec:uncertainty_resolving_B}
Denote the local-map frame of the odometry as $\{W\}$. An ideal point-to-plane ICP problem is given by 
\begin{equation}
	\underset{{^{W}_{I}\mathbf{R}},{^{W}_{I}\mathbf{t}}}{\min}\sum_{i=1}^n{\left\| \mathbf{u}_{i}^{\mathrm{T}}\left( {^{W}_{I}\mathbf{R}}{^I\mathbf{p}_i}+{^{W}_{I}\mathbf{t}}-\mathbf{q}_i \right) \right\| ^2}
\end{equation}
where ${^{W}_{I}{\mathbf{R}}}\in \mathrm{SO}\left( 3 \right) $ and $^{W}_{I}\mathbf{t}\in \mathbb{R} ^3$ are the decision variables, ${^I\mathbf{p}_i}$ is the ground-truth LiDAR point, $\mathbf{u}_{i}$ indicates the normal vector of the corresponding plane (obtained via the least squares method) from the map, and $\mathbf{q}_i$ indicates a point on the plane. This ICP problem can be solved using the Gauss-Newton or Levenberg–Marquardt method iteratively and incrementally. When the algorithm converges for the global minimizer, the last increment is given by solving the following optimization problem:
\begin{equation}
	\delta \boldsymbol{\zeta }^*=\mathbf 0=\underset{\delta \boldsymbol{\zeta }}{\mathrm{arg}\min}\sum_{i=1}^n{\left\| \varepsilon _i \right\| ^2}
	\label{eq:last_optimization}
\end{equation}
\begin{equation}
	\varepsilon_i=\mathbf{u}_{i}^{\mathrm{T}}\left(
	{^{W}_{I}\tilde{\mathbf{R}}^{*}}\mathrm{Exp}(\delta\boldsymbol\theta){^I\mathbf{p}_{i}}
	+{^{W}_{I}\tilde{\mathbf{t}}^{*}}+\delta\mathbf{t}-\mathbf{q}_{i}
	\right)
\end{equation}
\begin{equation}
	\delta\boldsymbol\zeta=[\delta\mathbf{t}^{\mathrm{T}},\delta\boldsymbol\theta^{\mathrm{T}}]^{\mathrm{T}}
	\label{eq:icp_product_retraction}
\end{equation}
where $\mathrm{Exp}( \cdot ) =\exp ( ( \cdot ) ^{\land} ) $ , $\delta \mathbf{t}\in \mathbb{R} ^3$ and $\delta \boldsymbol{\theta }\in \mathfrak{so} \left( 3 \right) $ compose the calculated increment in this iteration,  $^{W}_{\ I}{\tilde{\mathbf{R}}}^*\in \mathrm{SO}(3)$ and $^{W}_{\ I}{\tilde{\mathbf{t}}}^*\in \mathbb{R} ^3$ compose the optimal solution from the second last iteration. 

According to the implicit function theorem \cite{censi_2007} and \eqref{eq:final_point}, there is an implicit map $\mathbf f$ from the LiDAR points to the final increment:
\begin{equation}
	\begin{aligned}
		\delta\boldsymbol\zeta&=\mathbf{f}\left( ^I\mathbf{p}_{1:n} \right)
		\\
		&=\mathbf{f}\left( ^I\tilde{\mathbf{p}}_{1:n} \right) +\sum_{i=1}^n{\frac{\partial \mathbf{f}\left( ^I\tilde{\mathbf{p}}_{1:n} \right)}{\partial ^I\mathbf{p}_i}\left( ^I\mathbf{p}_i-{^I\tilde{\mathbf{p}}_i} \right)}+{\mathbf{r}^{\mathrm{nl}}_{\mathrm{icp}}}
	\end{aligned}
	\label{eq:d_zeta_eq_f}
\end{equation}
with
\begin{equation}
	\mathbf{f}\left( ^I\tilde{\mathbf{p}}_{1:n} \right) =\delta\boldsymbol\zeta^*= \mathbf{0} 
\end{equation}
\begin{equation}
	\begin{aligned}
		&\frac{\partial \mathbf{f}\left( ^I\tilde{\mathbf{p}}_{1:n} \right)}{\partial ^I\mathbf{p}_i}
		\\
		&=-\left( \frac{\partial ^2J\left( \delta\boldsymbol\zeta^*,^I\tilde{\mathbf{p}}_{1:n} \right)}{\partial \delta\boldsymbol\zeta^2} \right) ^{-1}
		\frac{\partial ^2J\left( \delta\boldsymbol\zeta^*,^I\tilde{\mathbf{p}}_{1:n} \right)}{\partial ^I\mathbf{p}_i\partial \delta\boldsymbol\zeta}
		\\
		&=-\left( \sum_{j=1}^n{\mathbf{J}_{\varepsilon,j}^{\mathrm{T}}\mathbf{J}_{\varepsilon,j}} \right) ^{-1} \left[ \begin{array}{c}
			\mathbf{u}_{i}\mathbf{B}_{i}\\
			{^I\tilde{\mathbf{p}}_i^{\land}}\mathbf{B}_{i}^{\mathrm{T}}\mathbf{B}_{i}\\
		\end{array} \right] 
	\end{aligned}
	\label{eq:d_f_d_p_original}
\end{equation}
where $\mathbf{B}_{i}=\mathbf{u}_{i}^{\mathrm{T}}{^{W}_{I}\tilde{\mathbf{R}}^{*}}$, $\mathbf{J}_{\varepsilon,i}=[\mathbf{u}_{i}^{\mathrm{T}},-\mathbf{B}_{i}{^I\tilde{\mathbf{p}}_i}^{\land}]$, and $\mathbf{r}^{\mathrm{nl}}_{\mathrm{icp}}$ contains the neglected higher-order terms. The derivation is given in Appendix~\ref{sec:appd2}. Based on \eqref{eq:d_zeta_eq_f} and the closed-form Jacobian matrix above, the ellipsoidal set that contains the last increment can be given via the Minkowski sum operation (see Section \ref{sec:minkoeski}):
\begin{equation}
	\begin{aligned}
		\delta\boldsymbol\zeta=&\delta\boldsymbol\zeta-\delta\boldsymbol\zeta^*\in \mathcal{E} \left( \mathbf{0},\tilde{\mathbf{Q}}^{\zeta} \right) 
		\\
		\triangleq &\left( \bigoplus_{i=1}^n{}_{\mathcal{E}}\mathcal{E} \left( \mathbf{0},\frac{\partial \mathbf{f}\left( ^I\tilde{\mathbf{p}}_{1:n} \right)}{\partial ^I\mathbf{p}_i}{^I\mathbf{P}_{i}^{\mathrm{p}}}\frac{\partial \mathbf{f}\left( ^I\tilde{\mathbf{p}}_{1:n} \right)}{\partial ^I\mathbf{p}_i}^{\mathrm{T}} \right) \right) 
		\\
		&\oplus_{\mathcal{E}}\mathcal{E} \left( \mathbf 0,{\mathbf{P}^{\mathrm{nl}}_{\mathrm{icp}}} \right)
	\end{aligned}
\end{equation}
where $\mathcal{E} \left( \mathbf{0},\mathbf{P}^\mathrm{nl}_{\mathrm{icp}} \right) $ bounds $\mathbf{r}^{\mathrm{nl}}_{\mathrm{icp}}$, and $\mathbf{P}^\mathrm{nl}_{\mathrm{icp}}$ is calibrated by the Monte Carlo sampling \cite{cong_2022}. Eventually, the estimated pose from the point-to-plane ICP algorithm with UBB uncertainty is
\begin{equation}
	_{I}^{W}\tilde{\mathbf{R}}={_{I}^{W}\tilde{\mathbf{R}}}^*\mathrm{Exp(}\delta \boldsymbol{\theta }), \quad {_{I}^{W}\tilde{\mathbf{t}}}={_{I}^{W}\tilde{\mathbf{t}}}^*+\delta \mathbf{t}
\end{equation}
\begin{equation}
	\delta\boldsymbol\zeta=[\delta\mathbf{t}^{\mathrm{T}},\delta\boldsymbol\theta^{\mathrm{T}}]^{\mathrm{T}}\in\mathcal E(\mathbf0,\tilde{\mathbf Q}^{\zeta})
\end{equation}

\section{On-Manifold Ellipsoidal Set-Membership Filter}
\label{sec:om_smf}
\subsection{System State}
The system state to be estimated at time $k$ in our system is
\begin{equation}
	\mathbf{x}_k=\left[ \mathbf{t}_{k}^{\mathrm{T}},\mathbf{v}_{k}^{\mathrm{T}},\mathbf{R}_{k}^{\mathrm{T}} \right] ^{\mathrm{T}}\in \mathcal{M} ,\ 	\mathcal{M} \triangleq \mathbb{R} ^6\times \mathrm{SO}\left( 3 \right) 
\end{equation}
where $\mathbf{t}_{k} \in \mathbb{R}^3$ is the translation, $\mathbf{v}_{k} \in \mathbb{R}^3$ is the velocity, and $\mathbf{R}_{k} \in \mathrm{SO(3)}$ is the rotation matrix. The system state is the description of $\{I\}$ relative to $\{W\}$, where $m$ is the local-map index. For simplicity, the reference frame of the state will be omitted in this section. 

To implement an on-manifold set-membership filter, it is essential to separate the nominal state and the error state. In our system, the nominal state is defined as
\begin{equation}
	\hat{\mathbf{x}}_k=\left[ \hat{\mathbf{t}}_{k}^{\mathrm{T}},\hat{\mathbf{v}}_{k}^{\mathrm{T}},\hat{\mathbf{R}}_{k}^{\mathrm{T}} \right] ^{\mathrm{T}}
	\label{eq:nominal_state}
\end{equation}
and the error state is defined as
\begin{equation}
	\delta \hat{\mathbf{x}}_k=\mathbf{x}_k\boxminus _{\mathcal{M}}\hat{\mathbf{x}}_k=\left[ \delta \hat{\mathbf{t}}_{k}^{\mathrm{T}},\delta \hat{\mathbf{v}}_{k}^{\mathrm{T}},\delta \hat{\boldsymbol{\theta}}_{k}^{\mathrm{T}} \right] ^{\mathrm{T}}
	\label{eq:error_state}
\end{equation}
where $\boxminus _{\mathcal{M}}$ is the encapsulated minus operator \cite{fast_lio2} for $\mathcal{M}$, $\delta \hat{\mathbf{t}}_{k}^{}\in \mathbb{R} ^3$, $\delta \hat{\mathbf{v}}_{k}^{}\in \mathbb{R} ^3$, and $\delta \hat{\boldsymbol{\theta}}_{k}^{}\in \mathfrak{so}\left( 3 \right) $. The error state in our system is assumed to be UBB and be within separated ellipsoidal sets:
\begin{equation}
	\delta \hat{\mathbf{t}}_k\in \mathcal{E} \left( \mathbf 0,\hat{\mathbf{P}}_{k}^{\mathrm{t}} \right) ,\delta \hat{\mathbf{v}}_k\in \mathcal{E} \left( \mathbf 0,\hat{\mathbf{P}}_{k}^{\mathrm{v}} \right) ,\delta \hat{\boldsymbol{\theta}}_k\in \mathcal{E} \left( \mathbf 0,\hat{\mathbf{P}}_{k}^{\theta} \right) 
\end{equation}
where $\mathcal{E} ( \mathbf{0},\hat{\mathbf{P}}_{k}^{\mathrm{t}} )$ is considered as the translational \emph{local protection level}, and $\mathcal{E} ( \mathbf{0},\hat{\mathbf{P}}_{k}^{\theta} )$ is considered as the rotational \emph{local protection level}.

Due to the complexity of the SMF, and in order to reduce the dimensionality of the system state and limit the conservativeness of the SMF, we do not include IMU biases as part of the system state. Fortunately, based on the UBB assumption, our state estimation process can account for time-varying biases and reflect their uncertainty in the state estimation results.

\subsection{Prediction}
Based on the kinematic model \cite{fast_lio2} of the system state, the motion model of the nominal state \eqref{eq:nominal_state} is
\begin{equation}
	\begin{aligned}
		\check{\mathbf{t}}_{k+1}&=\hat{\mathbf{t}}_k+\hat{\mathbf{v}}_k\Delta t+\frac{1}{2}\left( \hat{\mathbf{R}}_k(\tilde{\mathbf{a}}_k-\mathbf{b}_{\mathrm{a}}) \right) \Delta t^2+\frac{1}{2}\mathbf{g}\Delta t^2
		\\
		\check{\mathbf{v}}_{k+1}&=\hat{\mathbf{v}}_k+\hat{\mathbf{R}}_k(\tilde{\mathbf{a}}_k-\mathbf{b}_{\mathrm{a}})\Delta t+\mathbf{g}\Delta t
		\\
		\check{\mathbf{R}}_{k+1}&=\hat{\mathbf{R}}_k\mathrm{Exp}\left( (\tilde{\boldsymbol{\omega}}_k-\mathbf{b}_{\mathrm{g}})\Delta t \right) 
	\end{aligned}
	\label{eq:nominal_motion_model}
\end{equation}
where $\Delta t$ is the sampling interval, $\tilde{\mathbf{a}}_k$ and $\tilde{\boldsymbol{\omega}}_k$ are the IMU measurements, $\mathbf{b}_{\mathrm{a}}$ and $\mathbf{b}_{\mathrm{g}}$ are the IMU biases (obtained using the IMU measurements during the static initialization before moving), $\mathbf{g}$ is the gravity. In the prediction step, the nominal state is predicted via \eqref{eq:nominal_motion_model} without considering any noise.

According to the definition of the error state \eqref{eq:error_state}, its motion model is
\begin{equation}
	\begin{aligned}
		\delta \check{\mathbf{t}}_{k+1}=&\delta \hat{\mathbf{t}}_k+\delta \hat{\mathbf{v}}_k\Delta t
		\\
		\delta \check{\mathbf{v}}_{k+1}=&\delta \hat{\mathbf{v}}_k-\hat{\mathbf{R}}_k(\tilde{\mathbf{a}}_k-\mathbf{b}_{\mathrm{a}})^{\land}\delta \hat{\boldsymbol{\theta}}_{k}\Delta t-\hat{\mathbf{R}}_k\delta \mathbf{b}_{\mathrm{a}}\Delta t-\boldsymbol{\eta }_{\mathrm{v},k}
		\\
		\delta \check{\boldsymbol{\theta}}_{k+1}=&\mathrm{Exp}\left( -(\tilde{\boldsymbol{\omega}}_k-\mathbf{b}_{\mathrm{g}})\Delta t \right) \delta \hat{\boldsymbol{\theta}}_k-\delta \mathbf{b}_{\mathrm{g}}\Delta t-\boldsymbol{\eta }_{\theta ,k}
	\end{aligned}
\end{equation}
where $\delta \mathbf{b}_{\textrm{a}}$ and $\delta \mathbf{b}_{\textrm{g}}$ are the error of the calibrated IMU biases, and $\boldsymbol{\eta }_{\mathrm{v},k}$ and $\boldsymbol{\eta }_{\theta,k}$ are the noises. In our system, the errors of biases and noises are assumed to be UBB and within ellipsoidal sets:
\begin{equation}
	\delta \mathbf{b}_{\mathrm{a}}\in \mathcal{E} \left( \mathbf{0},\mathbf{P}^{\mathrm{ba}} \right) ,\ \delta \mathbf{b}_{\mathrm{g}}\in \mathcal{E} \left( \mathbf{0},\mathbf{P}^{\mathrm{bg}} \right) 
\end{equation}
\begin{equation}
	\boldsymbol{\eta }_{\mathrm{v},k}\in \mathcal{E} \left( 0,\Delta {t}^2\mathbf{N}_{k}^{\mathrm{a}} \right) ,\ \boldsymbol{\eta }_{\theta ,k}\in \mathcal{E} \left( 0,\Delta {t}^2\mathbf{N}_{k}^{\mathrm{g}} \right) 
\end{equation}
\begin{equation}
	\mathbf{N}_{k}^{\mathrm{a}}=\mathrm{diag}\left( 3b_{a}^{2},3b_{a}^{2},3b_{a}^{2} \right) ,\ \mathbf{N}_{k}^{\mathrm{g}}=\mathrm{diag}\left( 3b_{g}^{2},3b_{g}^{2},3b_{g}^{2} \right) 
\end{equation}
In the prediction stage, based on the motion model, the ellipsoidal sets that contain the predicted error state can be calculated as follows:
\begin{equation}
	\delta \check{\mathbf{t}}_{k+1}\in \mathcal{E} \left( \mathbf0,\check{\mathbf{P}}_{k+1}^{\mathrm{t}} \right) \triangleq \mathcal{E} \left( \mathbf0,\hat{\mathbf{P}}_{k}^{\mathrm{t}} \right)\oplus _{\mathcal{E}} \mathcal{E} \left( \mathbf0,\Delta t^2\hat{\mathbf{P}}_{k}^{\mathrm{v}} \right) 
	\label{eq:predicted_error_t}
\end{equation}
\begin{equation}
	\begin{aligned}
		\delta \check{\mathbf{v}}_{k+1}\in \mathcal{E} \left( \mathbf0,\check{\mathbf{P}}_{k+1}^{\mathrm{v}} \right)& \triangleq  \mathcal{E} \left( \mathbf0,\hat{\mathbf{P}}_{k}^{\mathrm{v}} \right) \oplus_{\mathcal{E}} \mathcal{E} \left( \mathbf0,\mathbf{C}_k\hat{\mathbf{P}}_{k}^{\theta}\mathbf{C}_{k}^{\mathrm{T}} \right) 
		\\
		\oplus& _{\mathcal{E}}\mathcal{E} \left( \mathbf0,\mathbf{D}_k\mathbf{P}^{\mathrm{ba}}\mathbf{D}_{k}^{\mathrm{T}} \right) \oplus _{\mathcal{E}}\mathcal{E} \left( \mathbf0,\Delta {t}^2\mathbf{N}_{k}^{\mathrm{a}} \right) 
	\end{aligned}
	\label{eq:predicted_error_v}
\end{equation}
\begin{equation}
	\begin{aligned}
		\delta \check{\boldsymbol{\theta}}_{k+1}\in \mathcal{E} \left( \mathbf0,\check{\mathbf{P}}_{k+1}^{\theta} \right) \triangleq &\mathcal{E} \left( \mathbf0,\mathbf{E}_k\hat{\mathbf{P}}_{k}^{\theta}\mathbf{E}_{k}^{\mathrm{T}} \right) \oplus_{\mathcal{E}} \mathcal{E} \left( \mathbf0,\Delta {t}^2\mathbf{P}^{\mathrm{bg}} \right) 
		\\
		&\oplus_{\mathcal{E}} \mathcal{E} \left( \mathbf0,\Delta {t}^2\mathbf{N}_{k}^\mathrm{g} \right) 
	\end{aligned}
	\label{eq:predicted_error_theta}
\end{equation}
where $\mathbf{C}_k=-\hat{\mathbf{R}}_k(\tilde{\mathbf{a}}_k- \mathbf{b}_{\mathrm{a}})^{\land}\Delta {t}$, $\mathbf{D}_k=-\hat{\mathbf{R}}_k\Delta {t}$, and $\mathbf{E}_k=\mathrm{Exp}\left( -(\tilde{\boldsymbol{\omega}}_k- \mathbf{b}_{\mathrm{g}} )\Delta t \right)$.

\subsection{Update}
\label{sec:update}
In the update stage, the ICP pose set from Section~\ref{sec:uncertainty_resolving} is used directly as the observation. At time $k+1$, we have
\begin{equation}
	_{I}^{W}\tilde{\mathbf{R}}_{k+1}={_{I}^{W}\tilde{\mathbf{R}}_{k+1}^{*}}\mathrm{Exp(}\delta \boldsymbol{\theta }_{k+1}),\quad {_{I}^{W}\tilde{\mathbf{t}}_{k+1}}={_{I}^{W}\tilde{\mathbf{t}}_{k+1}^{*}}+\delta \mathbf{t}_{k+1}
\end{equation}
\begin{equation}
	\delta \boldsymbol\zeta_{k+1}
	=[\delta\mathbf t_{k+1}^{\mathrm T},\delta\boldsymbol\theta_{k+1}^{\mathrm T}]^{\mathrm T}
	\in\mathcal E(\mathbf0,\tilde{\mathbf Q}_{k+1}^{\zeta})
	\label{eq:icp_pose_product_set}
\end{equation}
Let $\mathbf F^{\mathrm t}=[\mathbf I_{3\times3}\ \mathbf O_{3\times3}]$, the translational marginal of the ICP pose set is
\begin{equation}
	\tilde{\mathbf t}_{k+1}
	\in\mathcal E(\tilde{\mathbf t}_{k+1}^{*},\tilde{\mathbf Q}_{k+1}^{\mathrm t}),
	\quad
	\tilde{\mathbf Q}_{k+1}^{\mathrm t}
	=\mathbf F^{\mathrm t}\tilde{\mathbf Q}_{k+1}^{\zeta}(\mathbf F^{\mathrm t})^{\mathrm T}
	\label{eq:icp_additive_translation_set}
\end{equation}
Grounded in the relationship between the observed translation and the predicted nominal translation, the observed additive translation error is
\begin{equation}
	\delta \tilde{\mathbf{t}}_{k+1}\in \mathcal{E} \left( \tilde{\mathbf{t}}_{k+1}^{*}-\check{\mathbf{t}}_{k+1},\tilde{\mathbf{Q}}_{k+1}^{\mathrm{t}} \right)  
	\label{eq:observed_error_t}
\end{equation}
Since the predicted ellipsoidal set \eqref{eq:predicted_error_t} and the observed ellipsoidal set \eqref{eq:observed_error_t} both contain the unknown ground-truth translation, their intersection (see Section \ref{sec:intersection}) is a tighter set that also contains the ground-truth translation, which is the outcome of fusing the IMU measurements and the LiDAR measurements. Hence, the predicted error translation state can be updated as follows:
\begin{equation}
	\begin{aligned}
		\delta \hat{\mathbf{t}}_{k+1}\in \ & \mathcal{E} \left( \delta \hat{\mathbf{t}}_{k+1}^{+},\hat{\mathbf{P}}_{k+1}^{\mathrm{t}} \right)
		\\
		&\triangleq  \mathcal{E} \left( \mathbf0,\check{\mathbf{P}}_{k+1}^{\mathrm{t}} \right) \cap_\mathcal{E} \mathcal{E} \left( \tilde{\mathbf{t}}_{k+1}^{*}-\check{\mathbf{t}}_{k+1},\tilde{\mathbf{Q}}_{k+1}^{\mathrm{t}} \right) 
	\end{aligned}
	\label{eq:updated_error_t}
\end{equation}

In terms of the velocity, by leveraging Euler's method, the connection between the observed translation and the observed velocity is given as $\tilde{\mathbf{t}}_{k+1}-\tilde{\mathbf{t}}_k=\tilde{\mathbf{v}}_{k+1}\Delta t$. Consequently, the observed velocity satisfies
\begin{equation}
	\tilde{\mathbf{v}}_{k+1}\Delta t\in \mathcal{E} (\Delta \tilde{\mathbf{t}}_{k+1},\tilde{\mathbf{Q}}_{k+1}^{\Delta \mathrm{t}})\triangleq \mathcal{E} (\tilde{\mathbf{t}}_{k+1}^{*},\tilde{\mathbf{Q}}_{k+1}^{\mathrm{t}})\oplus _{\mathcal{E}}\mathcal{E} (-\tilde{\mathbf{t}}_{k}^{*},\tilde{\mathbf{Q}}_{k}^{\mathrm{t}})
\end{equation}
\begin{equation}
	\tilde{\mathbf{v}}_{k+1}\in \mathcal{E} \left( \frac{1}{\Delta t}\Delta \tilde{\mathbf{t}}_{k+1},\frac{1}{\Delta t^2}\tilde{\mathbf{Q}}_{k+1}^{\Delta \mathrm{t}} \right) 
\end{equation}
And the ellipsoidal set that contains the observed error velocity is
\begin{equation}
	\delta \tilde{\mathbf{v}}_{k+1}\in \mathcal{E} \left( \frac{1}{\Delta t}\Delta \tilde{\mathbf{t}}_{k+1}-\check{\mathbf{v}}_{k+1},\frac{1}{\Delta t^2}\tilde{\mathbf{Q}}_{k+1}^{\Delta \mathrm{t}} \right) 
\end{equation}
Thus, the predicted error velocity state can be updated as follows:
\begin{equation}
	\begin{aligned}
		\delta \hat{\mathbf{v}}_{k+1}\in \ & \mathcal{E} \left( \delta \hat{\mathbf{v}}_{k+1}^{+},\hat{\mathbf{P}}_{k+1}^{\mathrm{v}} \right) 
		\\
		&\triangleq \mathcal{E} ( \mathbf{0},\check{\mathbf{P}}_{k+1}^{\mathrm{v}} ) \cap_\mathcal{E} \mathcal{E} ( \frac{1}{\Delta t}\Delta \tilde{\mathbf{t}}_{k+1}-\check{\mathbf{v}}_{k+1},\frac{1}{\Delta t^2}\tilde{\mathbf{Q}}_{k+1}^{\Delta \mathrm{t}} ) 
	\end{aligned}
	\label{eq:updated_error_v}
\end{equation}

Let $\mathbf F^{\mathrm r}=[\mathbf O_{3\times3}\ \mathbf I_{3\times3}]$, the rotational marginal of the ICP pose set is
\begin{equation}
	\tilde{\mathbf{R}}_{k+1}=\tilde{\mathbf{R}}_{k+1}^{*}\mathrm{Exp}\left( \delta \boldsymbol{\theta }_{k+1} \right) ,\quad \delta \boldsymbol{\theta }_{k+1}\in \mathcal{E} \left( \mathbf{0},\tilde{\mathbf{Q}}_{k+1}^{\mathrm{r}} \right) 
\end{equation}
\begin{equation}
	\tilde{\mathbf{Q}}_{k+1}^{\mathrm{r}}=\mathbf{F}^{\mathrm{r}}\tilde{\mathbf{Q}}_{k+1}^{\zeta}(\mathbf{F}^{\mathrm{r}})^{\mathrm{T}}
\end{equation}
The connection between the predicted nominal rotation state and the observed rotation is
\begin{equation}
	\tilde{\mathbf{R}}_{k+1}^{*}\mathrm{Exp}\left( \delta \boldsymbol{\theta }_{k+1} \right) =\check{\mathbf{R}}_{k+1}\mathrm{Exp}\left( \delta \tilde{\boldsymbol{\theta}}_{k+1} \right) 
\end{equation}
\begin{equation}
	\begin{aligned}
		&\delta \tilde{\boldsymbol{\theta}}_{k+1}= \mathrm{Log}\left( \check{\mathbf{R}}_{k+1}^{\mathrm{T}}\tilde{\mathbf{R}}_{k+1}^{*} \right) +{\mathbfcal{J}} ^{-1}\delta \boldsymbol{\theta }_{k+1} +\mathbf{r}_{\theta}^{\mathrm{nl}}
		\\
		&\in \mathcal{E} \left( \mathrm{Log}\left( \check{\mathbf{R}}_{k+1}^{\mathrm{T}}\tilde{\mathbf{R}}_{k+1}^{*} \right) ,\mathbfcal{J} ^{-1}\tilde{\mathbf{Q}}_{k+1}^{\mathrm{r}}\mathbfcal{J} ^{-\mathrm{T}} \right) \oplus _{\mathcal{E}}\mathcal{E} \left(\mathbf 0,\mathbf{P}_{\theta}^{\mathrm{nl}} \right) 
	\end{aligned}
\end{equation}
where $\mathrm{Log(}\cdot )=(\log\mathrm{(}\cdot ))^{\lor}$, $( \cdot ) ^{\lor}$ is the inverse of the skew-symmetric operator, and $\mathbfcal{J}$ is the right Jacobian \cite{barfoot_2017} of $\mathrm{Log}( \check{\mathbf{R}}_{k+1}^{\mathrm{T}}\tilde{\mathbf{R}}_{k+1}^{*} ) $, $\mathbf{r}_{\theta}^{\mathrm{nl}}\in \mathcal{E} ( 0,\mathbf{P}_{\theta}^{\mathrm{nl}})$ is the empirical compensation term for nonlinearity calibrated by the Monte Carlo sampling. Subsequently, the predicted error rotation state can be updated as follows:
\begin{equation}
	\begin{aligned}
		&\delta \hat{\boldsymbol{\theta}}_{k+1}\in \  \mathcal{E} \left( \delta \boldsymbol{\hat\theta }_{k+1}^{+},\hat{\mathbf{P}}_{k+1}^{\theta} \right) \triangleq\mathcal{E} \left( \mathbf{0},\check{\mathbf{P}}_{k+1}^{\theta} \right) 
		\\
		&\cap_\mathcal{E} (\mathcal{E} \left( \mathrm{Log}( \check{\mathbf{R}}_{k+1}^{\mathrm{T}}\tilde{\mathbf{R}}_{k+1}^{*} ) ,\mathbfcal{J} ^{-1}\tilde{\mathbf{Q}}_{k+1}^{\mathrm{r}}\mathbfcal{J} ^{-\mathrm{T}} )  \oplus _{\mathcal{E}}\mathcal{E} \left( 0,\mathbf{P}_{\theta}^{\mathrm{nl}} \right) \right)
	\end{aligned}
	\label{eq:updated_error_theta}
\end{equation}

Once the error state is updated, the nominal state can be updated via the error state. It yields
\begin{equation}
	\begin{aligned}
		\hat{\mathbf{t}}_{k+1}&=\check{\mathbf{t}}_{k+1}+\delta \hat{\mathbf{t}}_{k+1}^{+}
		\\
		\hat{\mathbf{v}}_{k+1}&=\check{\mathbf{v}}_{k+1}+\delta \hat{\mathbf{v}}_{k+1}^{+}
		\\
		\hat{\mathbf{R}}_{k+1}&=\check{\mathbf{R}}_{k+1}\mathrm{Exp}\left( \delta \hat{\boldsymbol{\theta}}_{k+1}^{+} \right) 
	\end{aligned}
	\label{eq:updated_nominal}
\end{equation}
Eventually, the centers of ellipsoidal sets of the error state are reset to zero. 

In practice, if the increment of the last iteration is significantly greater than the threshold, the ICP result is considered unreliable, and this update process will be ignored, and no map update will be carried out.

\begin{figure*}[htp]
	\centering
	\includegraphics[width=6.8in]{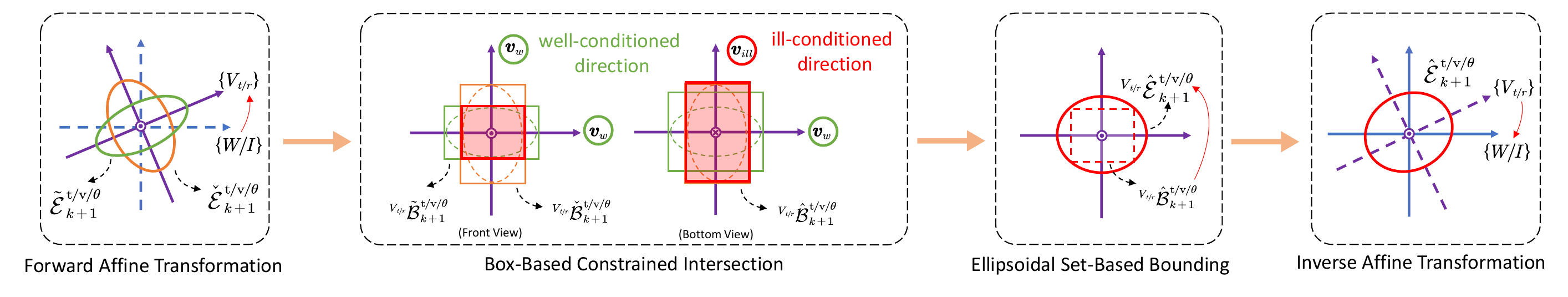}
	\caption{Overview of the proposed degeneracy management method. The forward affine transformation is conducted to obtain the predicted and observed ellipsoidal sets expressed in $\{V_{t/r}\}$. The box-based constrained intersection is designed to constrain the intersection operations of the ill-conditioned directions. The ellipsoidal set-based bounding is leveraged to maintain the shape of ellipsoidal sets. Finally, the inverse affine transformation is conducted to obtain the updated ellipsoidal sets expressed in $\{W\}$ or $\{I\}$.}
	\label{fig:degeneracy}
\end{figure*}

\subsection{Degeneracy Management}
For LiDAR-based localization algorithms, degeneracy indicates that the ICP problem is dominated by noises in under-constrained environments, and the algorithm converges to wrong local minimizers. In scenarios such as lawns and long tunnels, the degeneracy of LIO is inevitable. When the degeneracy manifests, the estimation from LIO will no longer be reliable. For our LIO with the protection level, it is extremely necessary to effectively handle the degeneracy and ensure the validity of the protection level.

\subsubsection{Degeneracy Detection}
We detect degeneracy directly from the eigenvalues of the Gauss-Newton Hessian of the point-to-plane ICP problem, which is
\begin{equation}
	\mathbf H=\frac{1}{n}\sum_{i=1}^{n}\mathbf J_{\varepsilon,i}^{\mathrm T}\mathbf J_{\varepsilon,i}
	=\begin{bmatrix}
		\mathbf H_{t} & \mathbf H_{tr}\\
		\mathbf H_{tr}^{\mathrm T} & \mathbf H_{r}
	\end{bmatrix}
\end{equation}
where $\mathbf H_{t}\in\mathbb R^{3\times3}$ and $\mathbf H_{r}\in\mathbb R^{3\times3}$ are the translational and rotational blocks, respectively. Since translation and rotation have different physical units, the two blocks are analyzed separately. Their eigendecompositions are
\begin{equation}
	\mathbf H_{t/r}=\mathbf V_{t/r}\boldsymbol\Lambda_{t/r}\mathbf V_{t/r}^{\mathrm T}, \quad
		\boldsymbol\Lambda_{t/r}=\mathrm{diag}\left(\lambda_{{t/r},1},\lambda_{{t/r},2},\lambda_{{t/r},3}\right)
\end{equation}
where $0\leq\lambda_{{t/r},1}\leq\lambda_{{t/r},2}\leq\lambda_{{t/r},3}$ and the columns $\mathbf v_{{t/r},j}$ of $\mathbf V_{t/r}$ are the corresponding eigenvectors. A small eigenvalue indicates low local curvature of the ICP objective along the corresponding eigenvector. Consequently, pose perturbations in this direction produce only a small increase in the objective value, implying that the direction is weakly constrained by the current geometry. We further define the eigenvalue ratio
\begin{equation}
	\rho _{t/r,j}=\lambda _{t/r,j}/\left( \lambda _{t/r,3}+\epsilon \right), \quad j=1,2,3
\end{equation}
where $\epsilon>0$ is a small constant for numerical stability. An absolute eigenvalue alone is affected by the scale of each block, whereas a ratio alone cannot identify the case in which all directions are weakly constrained. We therefore classify the direction $\mathbf v_{{t/r},j}$ as ill-conditioned if
\begin{equation}
	\lambda_{t/r,j}<\kappa_{t/r}^{\mathrm{abs}}
	\quad\text{or}\quad
	\rho_{t/r,j}<\kappa_{t/r}^{\mathrm{rel}}
	\label{eq:eigenvalue_degeneracy_test}
\end{equation}
where $\kappa_{t/r}^{\mathrm{abs}}$ and $\kappa_{t/r}^{\mathrm{rel}}$ are the absolute and relative eigenvalue thresholds, respectively. Degeneracy is declared when at least one translational or rotational direction satisfies \eqref{eq:eigenvalue_degeneracy_test}. The matrices $\mathbf V_{\mathrm t}$ and $\mathbf V_{\mathrm r}$, together with the detected ill-conditioned direction indices, are then passed to the degeneracy management procedure below. In this case, the standard error-state update in Section~\ref{sec:update} is replaced by the direction-constrained update, and the current point cloud is not inserted into the map.

\subsubsection{Degeneracy Management}
Denote the predicted error state as
$\delta \check{\mathbf{t}}_{k+1}\in \check{\mathcal{E}}_{k+1}^{\mathrm{t}}$, $\delta \check{\mathbf{v}}_{k+1}\in \check{\mathcal{E}}_{k+1}^{\mathrm{v}}$, $\delta \check{\boldsymbol{\theta}}_{k+1}\in \check{\mathcal{E}}_{k+1}^{\theta}$, the observed error state as $\delta \tilde{\mathbf{t}}_{k+1}\in \tilde{\mathcal{E}}_{k+1}^{\mathrm{t}}$, $\delta \tilde{\mathbf{v}}_{k+1}\in \tilde{\mathcal{E}}_{k+1}^{\mathrm{v}}$, $\delta \tilde{\boldsymbol{\theta}}_{k+1}\in \tilde{\mathcal{E}}_{k+1}^{\theta}$, and the updated state as $\delta \hat{\mathbf{t}}_{k+1}\in \hat{\mathcal{E}}_{k+1}^{\mathrm{t}}$, $\delta \hat{\mathbf{v}}_{k+1}\in \hat{\mathcal{E}}_{k+1}^{\mathrm{v}}$, $\delta \hat{\boldsymbol{\theta}}_{k+1}\in \hat{\mathcal{E}}_{k+1}^{\theta}$. If degeneracy is not detected, the updated error state is the intersection of the predicted error state and the observed error state. However, since there are ill-conditioned directions if degeneracy occurs, the update operation in these directions ought to be constrained. The overview of our degeneracy management method is depicted in Fig. \ref{fig:degeneracy}.

Denote the reference frame given by the eigenvectors of the translation part of the Hessian matrix as $\{V_t\}$, and the reference frame given by the eigenvectors of the rotation part of the Hessian matrix as $\{V_r\}$. One can obtain that the original ellipsoidal sets $\check{\mathcal{E}}_{k+1}^{\mathrm{t}}$,$\tilde{\mathcal{E}}_{k+1}^{\mathrm{t}}$,$\check{\mathcal{E}}_{k+1}^{\mathrm{v}}$ and $\tilde{\mathcal{E}}_{k+1}^{\mathrm{v}}$
that expressed in their original frame can be expressed in $\{V_t\}$ after the affine transformation with $\mathbf{V}_t^{\mathrm{T}}$, where $\mathbf{V}_t$ is comprised of the eigenvectors of the translation part of the Hessian matrix. Similarly, $\check{\mathcal{E}}_{k+1}^{\theta}$ and $\tilde{\mathcal{E}}_{k+1}^{\theta}$ can be expressed in $\{V_r\}$ with $\mathbf{V}_r^{\mathrm{T}}$, where $\mathbf{V}_r$ is comprised of the eigenvectors of the rotation part of the Hessian matrix. For simplicity and due to the consistency of the degeneracy management for different ellipsoidal sets, $\{{V}_{r/t}\}$ is used to describe the reference frame given by the eigenvectors, and $\check{\mathcal{E}}_{k+1}^{\mathrm{t}/\mathrm{v}/\theta}$, $\tilde{\mathcal{E}}_{k+1}^{\mathrm{t}/\mathrm{v}/\theta}$ and $\hat{\mathcal{E}}_{k+1}^{\mathrm{t}/\mathrm{v}/\theta}$ are utilized to denote the predicted, observed and updated ellipsoidal sets of the error state. The operation that obtains these sets is called the forward affine transformation.

After obtaining the form of ellipsoidal sets expressed in $\{{V}_{r/t}\}$, the box-based constrained intersection operation ought to be implemented to constrain the intersection in the ill-conditioned directions. The out-bounding boxes (see Section \ref{sec:box}) of the predicted and the observed ellipsoidal sets are both comprised of three intervals:
\begin{equation}
	^{V_{t/r}} \check{\mathcal{B}}  _{k+1}^{\mathrm{t}/\mathrm{v}/\theta}={^{V_{t/r}}\left[ \check{{b}}_1 \right] _{k+1}^{\mathrm{t}/\mathrm{v}/\theta}}\times ^{V_{t/r}}\left[ \check{{b}}_2 \right] _{k+1}^{\mathrm{t}/\mathrm{v}/\theta}\times ^{V_{t/r}}\left[ \check{{b}}_3 \right] _{k+1}^{\mathrm{t}/\mathrm{v}/\theta}
\end{equation}
\begin{equation}
	^{V_{t/r}} \tilde{\mathcal{B}}  _{k+1}^{\mathrm{t}/\mathrm{v}/\theta}=^{V_{t/r}}\left[ \tilde{{b}}_1 \right] _{k+1}^{\mathrm{t}/\mathrm{v}/\theta}\times ^{V_{t/r}}\left[ \tilde{{b}}_2 \right] _{k+1}^{\mathrm{t}/\mathrm{v}/\theta}\times ^{V_{t/r}}\left[ \tilde{{b}}_3 \right] _{k+1}^{\mathrm{t}/\mathrm{v}/\theta}
\end{equation}
If the direction is well-conditioned, the updated interval in this direction is the intersection of the corresponding predicted interval and the observed interval:
\begin{equation}
	^{V_{t/r}}\left[ \hat{{b}}_i \right] _{k+1}^{\mathrm{t}/\mathrm{v}/\theta}=^{V_{t/r}}\left[ \check{{b}}_i \right] _{k+1}^{\mathrm{t}/\mathrm{v}/\theta}\cap {^{V_{t/r}}\left[ \tilde{{b}}_i \right] _{k+1}^{\mathrm{t}/\mathrm{v}/\theta}}
\end{equation}
For ill-conditioned directions, the updated interval remains the predicted interval:
\begin{equation}
	{^{V_{t/r}}\left[ \hat{b}_{i} \right] _{k+1}^{\mathrm{t}/\mathrm{v}/\theta}}={^{V_{t/r}}\left[ \check{b}_{i} \right] _{k+1}^{\mathrm{t}/\mathrm{v}/\theta}}
\end{equation}
Finally, the updated box is composed as follows:
\begin{equation}
	^{V_{t/r}} \hat{\mathcal{B}} _{k+1}^{\mathrm{t}/\mathrm{v}/\theta}=^{V_{t/r}}\left[ \hat{b}_1 \right] _{k+1}^{\mathrm{t}/\mathrm{v}/\theta}\times ^{V_{t/r}}\left[ \hat{b}_2 \right] _{k+1}^{\mathrm{t}/\mathrm{v}/\theta}\times ^{V_{t/r}}\left[ \hat{b}_3 \right] _{k+1}^{\mathrm{t}/\mathrm{v}/\theta}
\end{equation}

Additionally, to maintain the ellipsoidal form of sets, the updated boxes need to be transformed back to ellipsoidal sets. With the purpose of establishing a set with greater conservatism, which can enhance the safety performance of the protection level, the out-bounding ellipsoidal set is formulated as the updated ellipsoidal set:
\begin{equation}
	^{V_{t/r}}\hat{\mathcal{E}}_{k+1}^{\mathrm{t}/\mathrm{v}/\theta}=\mathcal{E} \left( \left[ \begin{matrix}
		c_1&		c_2&		c_3\\
	\end{matrix} \right] ^{\mathrm{T}},\mathrm{diag}\left( 3r_{1}^{2},3r_{2}^{2},3r_{3}^{2} \right) \right) 
\end{equation}
where
\begin{equation}
	c_i=0.5\left( \mathrm{inf}\left( ^{V_{t/r}}\left[ \hat{b}_i \right] _{k+1}^{\mathrm{t}/\mathrm{v}/\theta} \right) +\mathrm{sup}\left( ^{V_{t/r}}\left[ \hat{b}_i \right] _{k+1}^{\mathrm{t}/\mathrm{v}/\theta} \right) \right) 
\end{equation}
\begin{equation}
	r_i=\mathrm{sup}\left( ^{V_{t/r}}\left[ \hat{b}_i \right] _{k+1}^{\mathrm{t}/\mathrm{v}/\theta} \right) -c_i
\end{equation}
Finally, the updated ellipsoidal set $\hat{\mathcal{E}}_{k+1}^{\mathrm{t}/\mathrm{v}/\theta}$ expressed in their original frames can be given through the inverse affine transformations with $\mathbf{V}_t$ and $\mathbf{V}_r$.

\section{Global Protection Level Propagation}
The uncertainty in Section \ref{sec:uncertainty_resolving} and Section \ref{sec:om_smf} is evaluated with respect to the active local map. During one local-map period, the map is treated as a fixed reference, and the posterior pose sets from the filter describe the \emph{local protection level}. Nevertheless, a local map is anchored using an uncertain historical pose. Neglecting this anchor uncertainty would make the reported bounds increasingly optimistic after successive local-map changes. We therefore maintain separate global translation and rotation sets that describe the accumulated uncertainty of the active local-map anchor and use them to convert the \emph{local protection levels} into \emph{global protection levels}.

Let $\{G\}$ denote the global world frame, and let $\{W\}$ denote the reference frame of local map with index $m$. The anchor pose with uncertainty of local map $m$ is represented by
\begin{equation}
	{}^{G}_{W}\mathbf{t}
	={}^{G}_{W}\hat{\mathbf{t}}+\delta\hat{\mathbf{t}}_{m}^{\mathrm{map}},
	\quad
	\delta\hat{\mathbf{t}}_{m}^{\mathrm{map}}
	\in\mathcal{E}\left(\mathbf{0},\hat{\mathbf{P}}_{m}^{\mathrm{t},\mathrm{map}}\right)
	\label{eq:map_translation_error}
\end{equation}
\begin{equation}
	{}^{G}_{W}\mathbf{R}
	={}^{G}_{W}\hat{\mathbf{R}}\mathrm{Exp}\left(\delta\hat{\boldsymbol\theta}_{m}^{\mathrm{map}}\right),
	\quad
	\delta\hat{\boldsymbol\theta}_{m}^{\mathrm{map}}
	\in\mathcal{E}\left(\mathbf{0},\hat{\mathbf{P}}_{m}^{\theta,\mathrm{map}}\right)
	\label{eq:map_rotation_error}
\end{equation}
For the first local map, it is aligned with $\{G\}$ and the shape matrices are set to the initialization uncertainty. A new local map $m+1$ with frame $\{W'\}$ is created when the displacement from the current local-map origin exceeds the map-switching threshold. The pose of $\{W'\}$ is initialized using the output pose from \emph{OM-ESMF} at the switching time $k=\mathrm{swc}$. Therefore, the estimated transformation from $\{W'\}$ to $\{W\}$ is given by ${}^{W}_{W'}\hat{\mathbf{R}}={}^{W}_{I}\hat{\mathbf{R}}_{\mathrm{swc}}$ and ${}^{W}_{W'}\hat{\mathbf{t}}={}^{W}_{I}\hat{\mathbf{t}}_{\mathrm{swc}}$. The uncertainties of them are the \emph{local protection levels} at the switching time. That is
\begin{equation}
	\delta\hat{\mathbf{t}}_{\mathrm{swc}}^{\mathrm{loc}}
	\in\mathcal{E}\left(\mathbf{0},\hat{\mathbf{P}}_{\mathrm{swc}}^{\mathrm{t}}\right),
	\quad
	\delta\hat{\boldsymbol\theta}_{\mathrm{swc}}^{\mathrm{loc}}
	\in\mathcal{E}\left(\mathbf{0},\hat{\mathbf{P}}_{\mathrm{swc}}^{\theta}\right)
	\label{eq:map_switch_local_sets}
\end{equation}
\begin{equation}
	\hat{\mathbf{P}}_{\mathrm{swc}}^{\mathrm{t}}
	=\left.\hat{\mathbf{P}}_{k}^{\mathrm{t}}\right|_{k=\mathrm{swc}},
	\quad
	\hat{\mathbf{P}}_{\mathrm{swc}}^{\theta}
	=\left.\hat{\mathbf{P}}_{k}^{\theta}\right|_{k=\mathrm{swc}}
	\label{eq:map_switch_shape_sources}
\end{equation}
The nominal anchor pose is updated according to
\begin{equation}
	{}^{G}_{W'}\hat{\mathbf{R}}
	={}^{G}_{W}\hat{\mathbf{R}}{}^{W}_{W'}\hat{\mathbf{R}},
	\quad
	{}^{G}_{W'}\hat{\mathbf{t}}
	={}^{G}_{W}\hat{\mathbf{t}}
	+{}^{G}_{W}\hat{\mathbf{R}}{}^{W}_{W'}\hat{\mathbf{t}}
	\label{eq:map_anchor_nominal_update}
\end{equation}
Linearizing the same composition with respect to the separate translation and rotation errors gives
\begin{equation}
	\delta\hat{\mathbf{t}}_{m+1}^{\mathrm{map}}
		={}\delta\hat{\mathbf{t}}_{m}^{\mathrm{map}}
		+{}^{G}_{W}\hat{\mathbf{R}}\delta\hat{\mathbf{t}}_{\mathrm{swc}}^{\mathrm{loc}}-{}^{G}_{W}\hat{\mathbf{R}}
		\left({}^{W}_{W'}\hat{\mathbf{t}}\right)^{\land}
		\delta\hat{\boldsymbol\theta}_{m}^{\mathrm{map}}
		+\mathbf{r}_{\mathrm{t},m}^{\mathrm{map},\mathrm{nl}}
	\label{eq:map_translation_propagation}
\end{equation}
\begin{equation}
	\delta\hat{\boldsymbol\theta}_{m+1}^{\mathrm{map}}
	=\left({}^{W}_{W'}\hat{\mathbf{R}}\right)^{\mathrm{T}}
	\delta\hat{\boldsymbol\theta}_{m}^{\mathrm{map}}
	+\delta\hat{\boldsymbol\theta}_{\mathrm{swc}}^{\mathrm{loc}}
	+\mathbf{r}_{\theta,m}^{\mathrm{map},\mathrm{nl}}
	\label{eq:map_rotation_propagation}
\end{equation}
The empirically calibrated remainder terms by the Monte Carlo sampling enclose the neglected higher-order components and satisfy
\begin{equation}
	\mathbf{r}_{\mathrm{t},m}^{\mathrm{map},\mathrm{nl}}
	\in\mathcal{E}\left(\mathbf{0},\mathbf{P}_{\mathrm{t},m}^{\mathrm{map},\mathrm{nl}}\right),
	\quad
	\mathbf{r}_{\theta,m}^{\mathrm{map},\mathrm{nl}}
	\in\mathcal{E}\left(\mathbf{0},\mathbf{P}_{\theta,m}^{\mathrm{map},\mathrm{nl}}\right)
\end{equation}
Consequently, the translational anchor uncertainty is propagated as
\begin{equation}
	\begin{aligned}
	\mathcal{E}\left(\mathbf{0},\hat{\mathbf{P}}_{m+1}^{\mathrm{t},\mathrm{map}}\right)
	&=\mathcal{E}\left(\mathbf{0},\hat{\mathbf{P}}_{m}^{\mathrm{t},\mathrm{map}}\right)\\
	\oplus&_{\mathcal{E}}
	\mathcal{E}\left(\mathbf{0},
	{}^{G}_{W}\hat{\mathbf{R}}
	\hat{\mathbf{P}}_{\mathrm{swc}}^{\mathrm{t}}
	({}^{G}_{W}\hat{\mathbf{R}})^{\mathrm{T}}\right)\\
	\oplus&_{\mathcal{E}}
	\mathcal{E}\left(\mathbf{0},
	{}^{G}_{W}\hat{\mathbf{R}}
	({}^{W}_{W'}\hat{\mathbf{t}})^{\land}
	\hat{\mathbf{P}}_{m}^{\theta,\mathrm{map}}
	(({}^{W}_{W'}\hat{\mathbf{t}})^{\land})^{\mathrm{T}}
	({}^{G}_{W}\hat{\mathbf{R}})^{\mathrm{T}}\right)\\
	\oplus&_{\mathcal{E}}
	\mathcal{E}\left(\mathbf{0},\mathbf{P}_{\mathrm{t},m}^{\mathrm{map},\mathrm{nl}}\right)
	\end{aligned}
	\label{eq:map_translation_shape_propagation}
\end{equation}
and the rotational anchor uncertainty is propagated as
\begin{equation}
	\begin{aligned}
	&\mathcal{E}\left(\mathbf{0},\hat{\mathbf{P}}_{m+1}^{\theta,\mathrm{map}}\right)\\
	&=\mathcal{E}\left(\mathbf{0},
	({}^{W}_{W'}\hat{\mathbf{R}})^{\mathrm{T}}
	\hat{\mathbf{P}}_{m}^{\theta,\mathrm{map}}
	{}^{W}_{W'}\hat{\mathbf{R}}\right)
	\oplus_{\mathcal{E}}
	\mathcal{E}\left(\mathbf{0},\hat{\mathbf{P}}_{\mathrm{swc}}^{\theta}\right)\\
	&\quad\oplus_{\mathcal{E}}
	\mathcal{E}\left(\mathbf{0},\mathbf{P}_{\theta,m}^{\mathrm{map},\mathrm{nl}}\right)
	\end{aligned}
	\label{eq:map_rotation_shape_propagation}
\end{equation}

At time $k$, the nominal global pose of the IMU is
\begin{equation}
	{}^{G}_{I}\hat{\mathbf{R}}_k
	={}^{G}_{W}\hat{\mathbf{R}}{}^{W}_{I}\hat{\mathbf{R}}_k,
	\quad
	{}^{G}_{I}\hat{\mathbf{t}}_k
	={}^{G}_{W}\hat{\mathbf{t}}
	+{}^{G}_{W}\hat{\mathbf{R}}{}^{W}_{I}\hat{\mathbf{t}}_k
	\label{eq:global_nominal_pose}
\end{equation}
The first-order global translation and rotation errors are, respectively,
\begin{equation}
	\delta\hat{\mathbf{t}}_{k}^{\mathrm{glb}}
		={}\delta\hat{\mathbf{t}}_{m}^{\mathrm{map}}
		+{}^{G}_{W}\hat{\mathbf{R}}\delta\hat{\mathbf{t}}_{k}^{\mathrm{loc}}-{}^{G}_{W}\hat{\mathbf{R}}
		({}^{W}_{I}\hat{\mathbf{t}}_k)^{\land}
		\delta\hat{\boldsymbol\theta}_{m}^{\mathrm{map}}
		+\mathbf{r}_{\mathrm{t},k}^{\mathrm{glb},\mathrm{nl}}
	\label{eq:global_translation_error}
\end{equation}
\begin{equation}
	\delta\hat{\boldsymbol\theta}_{k}^{\mathrm{glb}}
	=({}^{W}_{I}\hat{\mathbf{R}}_k)^{\mathrm{T}}
	\delta\hat{\boldsymbol\theta}_{m}^{\mathrm{map}}
	+\delta\hat{\boldsymbol\theta}_{k}^{\mathrm{loc}}
	+\mathbf{r}_{\theta,k}^{\mathrm{glb},\mathrm{nl}}
	\label{eq:global_rotation_error}
\end{equation}
where
\begin{equation}
	\mathbf{r}_{\mathrm{t},k}^{\mathrm{glb},\mathrm{nl}}
	\in\mathcal{E}\left(\mathbf{0},\mathbf{P}_{\mathrm{t},k}^{\mathrm{glb},\mathrm{nl}}\right),
	\quad
	\mathbf{r}_{\theta,k}^{\mathrm{glb},\mathrm{nl}}
	\in\mathcal{E}\left(\mathbf{0},\mathbf{P}_{\theta,k}^{\mathrm{glb},\mathrm{nl}}\right)
\end{equation}
and they are calibrated by the Monte Carlo sampling \cite{cong_2022}. Therefore, the translational \emph{global protection level} is
\begin{equation}
	\begin{aligned}
	&\delta\hat{\mathbf{t}}_{k}^{\mathrm{glb}}
	\in\mathcal{E}\left(\mathbf{0},\hat{\mathbf{P}}_{k}^{\mathrm{t},\mathrm{glb}}\right)\\
	&=\mathcal{E}\left(\mathbf{0},\hat{\mathbf{P}}_{m}^{\mathrm{t},\mathrm{map}}\right)
	\oplus_{\mathcal{E}}
	\mathcal{E}\left(\mathbf{0},
	{}^{G}_{W}\hat{\mathbf{R}}
	\hat{\mathbf{P}}_{k}^{\mathrm{t}}
	({}^{G}_{W}\hat{\mathbf{R}})^{\mathrm{T}}\right)\\
	&\quad\oplus_{\mathcal{E}}
	\mathcal{E}\left(\mathbf{0},
	{}^{G}_{W}\hat{\mathbf{R}}
	({}^{W}_{I}\hat{\mathbf{t}}_k)^{\land}
	\hat{\mathbf{P}}_{m}^{\theta,\mathrm{map}}
	(({}^{W}_{I}\hat{\mathbf{t}}_k)^{\land})^{\mathrm{T}}
	({}^{G}_{W}\hat{\mathbf{R}})^{\mathrm{T}}\right)\\
	&\quad\oplus_{\mathcal{E}}
	\mathcal{E}\left(\mathbf{0},\mathbf{P}_{\mathrm{t},k}^{\mathrm{glb},\mathrm{nl}}\right)
	\end{aligned}
	\label{eq:global_translation_pl}
\end{equation}
and the rotational \emph{global protection level} is
\begin{equation}
	\begin{aligned}
	&\delta\hat{\boldsymbol\theta}_{k}^{\mathrm{glb}}
	\in\mathcal{E}\left(\mathbf{0},\hat{\mathbf{P}}_{k}^{\theta,\mathrm{glb}}\right)\\
	&=\mathcal{E}\left(\mathbf{0},
	({}^{W}_{I}\hat{\mathbf{R}}_k)^{\mathrm{T}}
	\hat{\mathbf{P}}_{m}^{\theta,\mathrm{map}}
	{}^{W}_{I}\hat{\mathbf{R}}_k\right)
	\oplus_{\mathcal{E}}
	\mathcal{E}\left(\mathbf{0},\hat{\mathbf{P}}_{k}^{\theta}\right)\\
	&\quad\oplus_{\mathcal{E}}
	\mathcal{E}\left(\mathbf{0},\mathbf{P}_{\theta,k}^{\mathrm{glb},\mathrm{nl}}\right)
	\end{aligned}
	\label{eq:global_rotation_pl}
\end{equation}

\begin{figure}[tp]
	\centering
	\includegraphics[width=3.2in]{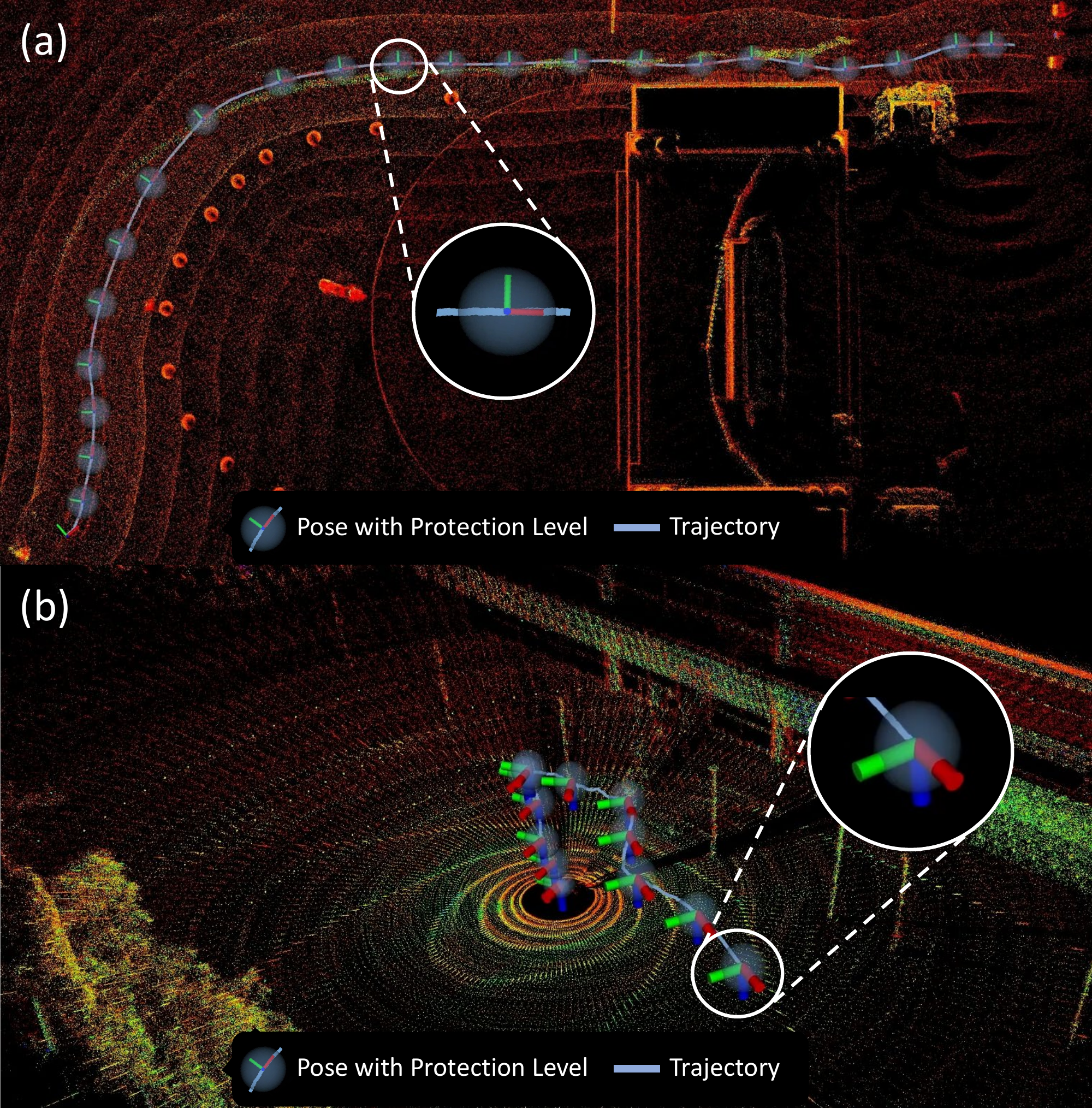}
	\caption{Estimated locations and protection levels from our system. Unlike other methods that only provide estimated locations, our system can provide deterministic protection levels. (a) \textit{gate\_02} in \textit{M2DGR}. (b) \textit{sbs\_03} in \textit{NTU VIRAL}.}
	\label{fig:m2dgr_result}
\end{figure}

\section{Experiments}
\subsection{Implementation}
In general, our system is implemented using C++ and the robot operating system (ROS). The ikd-Tree \cite{fast_lio2} is selected as the specific implementation of the map. To limit the computation burden of our system, the input point clouds are downsampled with the strategy given in \cite{fast_lio2}, and the maximum number of iterations for ICP is limited to 30. Furthermore, in order to fully leverage the performance advantages of the multi-core computing system, we employed the Intel thread building blocks (TBB) for parallel computing.

The UBB sensor-noise limits were specified independently of the test trajectories and their ground truth. For each sensor, initial bounds were obtained from the manufacturer's accuracy, resolution, bias, and operating-range specifications and were subsequently checked using separate stationary and repeated-measurement calibration data covering the operating ranges considered in the experiments. The LiDAR range bound $b_r$ accounts for range repeatability and calibration uncertainty, and the bearing bound $b_{\phi}$ accounts for angular resolution, beam-direction calibration, and residual deskewing error. The IMU bounds include accelerometer and gyroscope measurement errors and the admissible residual biases after initialization. To accommodate operating variations not represented in the calibration data, the largest calibration residual was enlarged by a fixed factor and the resulting bound was required to be no smaller than the corresponding manufacturer-specified maximum error. The same bounds were used for all sequences acquired with the same sensor configuration. Accordingly, the deterministic containment guarantee is conditional on the actual sensor errors remaining within these prescribed UBB limits; measurements outside the specified operating range or failing the sensor-validity checks are rejected.

\subsection{Experimental Setup}
In our experiments, FAST-LIO2 \cite{fast_lio2}, LIO-EKF \cite{lio_ekf}, LIO-SAM \cite{lio_sam}, DLIO \cite{chen_2023}, Brossard \textit{et al.} \cite{brossard_2020}, and PALoc \cite{paloc} are deployed as the SOTA methods to be compared with. FAST-LIO2, LIO-EKF, and Brossard \textit{et al.} provide covariance estimations, while PALoc obtains covariance estimations using the Laplace approximation. We used their 99.73\% confidence regions based on the chi-square distribution as protection levels. LIO-SAM and DLIO do not provide explicit uncertainty estimations.

\begin{figure*}[t]
	\centering
	\subfloat[\textit{gate\_01}]
	{\includegraphics[width=2.2in]{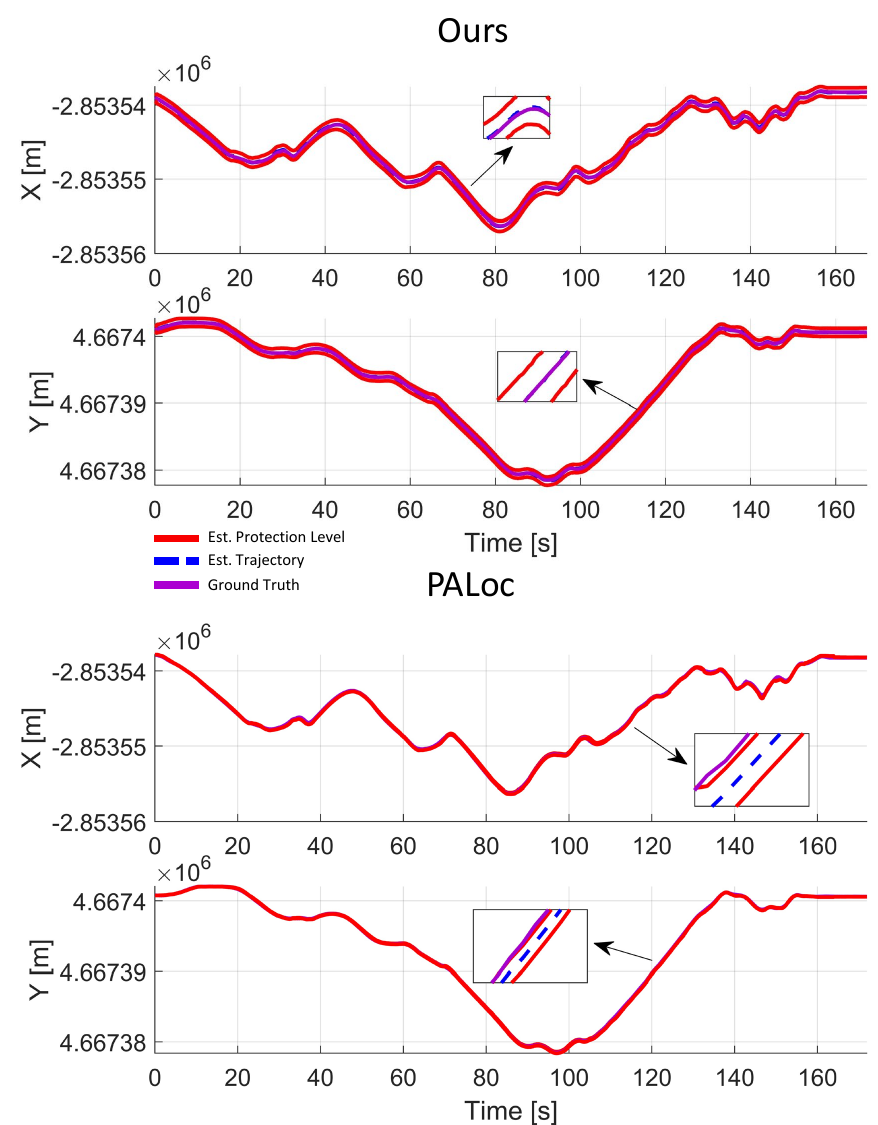}}
	\hspace{2px}
	\subfloat[\textit{hall\_01}]
	{\includegraphics[width=2.2in]{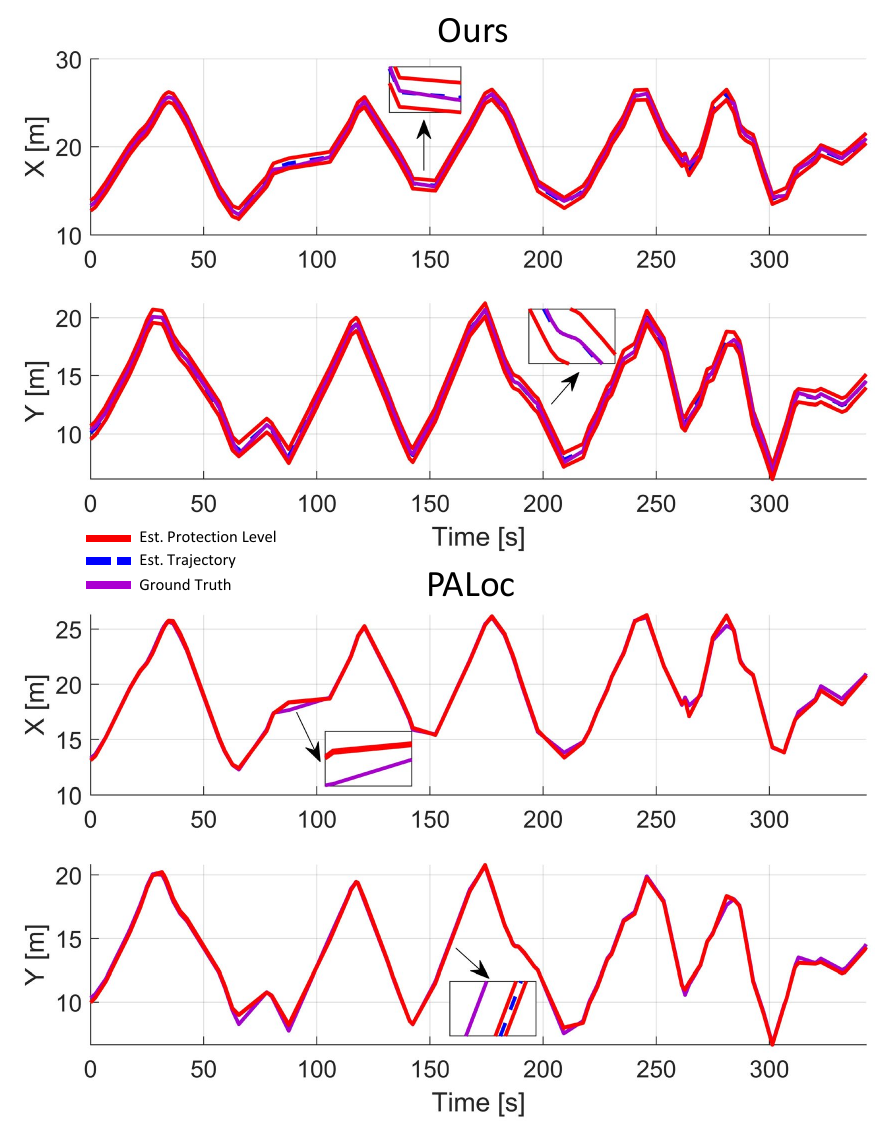}}
	\hspace{2px}
	\subfloat[\textit{room\_01}]
	{\includegraphics[width=2.2in]{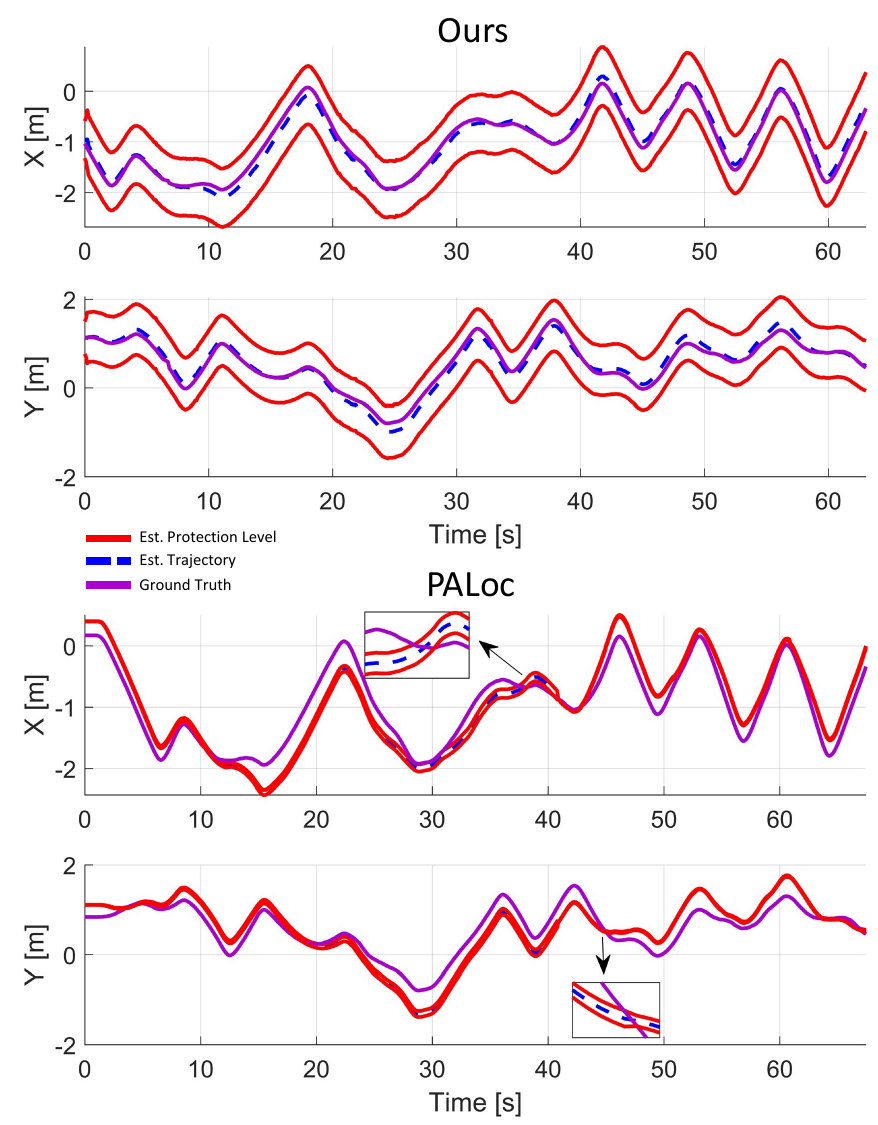}}
	\caption{Estimated \emph{local protection levels} and trajectories from our system and PALoc ($\sigma ^2=0.001$), and the ground truth using different sequences in \textit{M2DGR}. The protection levels estimated by PALoc cannot cover the ground truth inevitably, therefore, cannot effectively reflect the effectiveness of the navigation system. However, the protection levels obtained by our system can more effectively cover the ground truth.}
	\label{fig:bound_m2dgr}
\end{figure*}

\subsection{Datasets Results}
To verify the cross-platform performance of our system, different types of robots were chosen to conduct various experiments. We conducted thorough tests on our system using the following datasets with reliable ground truth obtained by 3D laser trackers or real-time kinematic GNSS:
\begin{enumerate}
	\item{\textit{M2DGR} \cite{zou_2022}: The M2DGR dataset was recorded with a wheeled mobile robot. In this dataset, Velodyne VLP-32C was deployed to capture 3D point clouds, and Handsfree A9 was used as the IMU.}
	
	\item{\textit{NTU VIRAL} \cite{xie_2022}: NTU VIRAL is a dataset recorded with an autonomous aerial vehicle. Ouster OS1-16 Gen11 and VectorNav VN100 were utilized as the sensors.}
	
	\item{\textit{SubT-MRS} \cite{subt-mrs}: SubT-MRS is an extremely challenging dataset with diverse sensor degradation scenarios. A high-precision 3D scanner was used to create ground-truth maps and trajectories.}
\end{enumerate}
It is noteworthy that in order to verify the effectiveness of our deterministic protection level, it is essential to conduct experiments using datasets and ground truth. In fact, our protection level can describe the reliability of the localization system in environments without ground truth.

The visualized pose estimations and protection levels using different robots are shown in Fig. \ref{fig:m2dgr_result}. Unlike other methods, in addition to providing the predicted trajectories, our system also offers deterministic protection levels, thereby enabling the online assessment of the safety and reliability of the localization system. Furthermore, it indicates that our system can construct detailed point cloud maps, which can serve for other downstream tasks such as motion planning.

To objectively assess the effectiveness of our system, the absolute trajectory error (ATE) is selected as the metric for evaluating the accuracy of localization. It is worth noting that the ellipsoidal sets estimated by our system are the feasible sets of the system state, and all values in one set are considered to be of equal status \cite{ehambram_2022}. Therefore, our system provides no optimal estimation result. However, to evaluate the accuracy of our system, the centers of the ellipsoidal sets were selected as the estimated values used for calculating ATE results in our experiments. The ATE results on the M2DGR dataset are shown in Table \ref{tab:acc_ate_m2dgr}, and the results on the NTU VIRAL dataset are shown in Table \ref{tab:acc_ate_ntuviral}. The results show that although our system was not constructed based on the optimal state estimation method, the estimated centers of the ellipsoidal sets by our system only have centimeter-level differences compared to the results estimated by the SOTA methods.

\begin{table}[tp]
	\caption{ATE [m] ($\downarrow$) on the M2DGR dataset.}
	\label{tab:acc_ate_m2dgr}
	\setlength{\tabcolsep}{1.9mm}
	\setlength{\extrarowheight}{0.6mm}
	\centering
	\begin{tabular}{lccccccc}
		\toprule
		& \rotatebox[origin=c]{90}{FAST-LIO2} & \rotatebox[origin=c]{90}{LIO-EKF} & \rotatebox[origin=c]{90}{LIO-SAM} & \rotatebox[origin=c]{90}{DLIO}
		& \rotatebox[origin=c]{90}{Brossard \it{et al.}} & \rotatebox[origin=c]{90}{PALoc} & \rotatebox[origin=c]{90}{Ours} \\
		\midrule
		street\_03 & 0.189 & 0.220 & 1.628 & 0.151 & 9.931 & 0.255 & \textbf{0.146} \\
		street\_08 & 0.205 & 0.366 & 5.319 & \textbf{0.135} & 21.271 & 0.256 & {0.153} \\
		gate\_01 & 0.172 & 0.564 & 0.148 & \textbf{0.134} & 4.436
		 & 0.177 & {0.139} \\
		gate\_03 & 0.213 & 0.227 & 0.111 & 0.117 & 8.086 & 0.210 & \textbf{0.095} \\
		hall\_01 & 0.291 & 0.326 & {0.243} & \textbf{0.183} & 3.013 & 0.303 & 0.247 \\
		hall\_02 & 0.530 & 0.547 & 0.538 & 0.349 & 1.268 & 0.524 & \textbf{0.267} \\
		hall\_03 & 0.547 & 0.631 & 0.658 & 0.466 & 2.089 & 0.550 & \textbf{0.370} \\
		room\_01 & 0.400 & 0.467 & 0.171 & 0.170 & 0.434 & 0.397 & \textbf{0.153} \\
		room\_02 & 0.315 & 0.476 & 0.125 & 0.209 & 0.534 & 0.316 & \textbf{0.124} \\
		room\_03 & 0.427 & 0.491 & \textbf{0.161} & 0.195 & 0.426 & 0.429 & 0.164 \\ \midrule
		Average & 0.329 & 0.432 & 0.910 & 0.211 & 5.149 & 0.342 & \textbf{0.186} \\
		\bottomrule
	\end{tabular}
\end{table}

\begin{table}[htp]
	\caption{ATE [m] ($\downarrow$) on the NTU VIRAL dataset.}
	\label{tab:acc_ate_ntuviral}
	\setlength{\tabcolsep}{1.9mm}
	\setlength{\extrarowheight}{0.6mm}
	\centering
	\begin{tabular}{lccccccc}
		\toprule
		& \rotatebox[origin=c]{90}{FAST-LIO2} & \rotatebox[origin=c]{90}{LIO-EKF} & \rotatebox[origin=c]{90}{LIO-SAM} & \rotatebox[origin=c]{90}{DLIO}
		& \rotatebox[origin=c]{90}{Brossard \it{et al.}} & \rotatebox[origin=c]{90}{PALoc} & \rotatebox[origin=c]{90}{Ours} \\
		\midrule
		eee\_01 & \textbf{0.069} & 0.245 & 0.075 & 0.148 & 13.542 & 0.097 & 0.120 \\
		eee\_02 & 0.083 & 0.114 & \textbf{0.069} & 0.143 & 16.797 & 0.183 & 0.117 \\
		eee\_03 & 0.111 & 0.142 & \textbf{0.101} & 0.197 & 13.280 & 0.112 & 0.173 \\
		nya\_01 &\textbf{ 0.053} & 0.150 & 0.076 & 0.130 & 10.800 & 0.117 & 0.118 \\
		nya\_02 & 0.090 & 0.133 & \textbf{0.089} & 0.160 & 5.394 & 0.191 & 0.146 \\
		nya\_03 & \textbf{0.109} & 0.129 & 0.143 & 0.178 & 8.756 & 0.251 & 0.153 \\
		sbs\_01 & \textbf{0.086} & 0.107 & 0.088 & 0.153 & 10.478 & 0.096 & 0.128 \\
		sbs\_02 & \textbf{0.075} & 0.150 & 0.084 & 0.135 & 17.625 & 0.103 & 0.120 \\
		sbs\_03 & \textbf{0.076} & 0.755 & 0.084 & 0.151 & 11.348 & 0.097 & 0.123 \\ \midrule
		Average & \textbf{0.084} & 0.214 & 0.090 & 0.155 & 12.002 & 0.139 & 0.133 \\
		\bottomrule
	\end{tabular}
\end{table}

Additionally, to objectively assess the estimated protection levels, the cover rate (CR) \cite{11081890} was chosen as the metric for evaluating the accuracy of the protection level, that is
\begin{equation}
	\mathrm{CR}_{\mathrm{trans}}=\frac{1}{K}\sum_{k=1}^K{\phi \left( \left( \mathbf{t}_k-\hat{\mathbf{t}}_k \right) ^{\mathrm{T}}\left( \hat{\mathbf{P}}_{k}^{\mathrm{t}} \right) ^{-1}\left( \mathbf{t}_k-\hat{\mathbf{t}}_k \right) \right)}	
\end{equation}
\begin{equation}
	\mathrm{CR}_{\mathrm{rot}}=\frac{1}{K}\sum_{k=1}^K{\phi \left( \mathrm{Log}( \hat{\mathbf{R}}_{k}^{\mathrm{T}}\mathbf{R}_k ) ^{\mathrm{T}}\left( \hat{\mathbf{P}}_{k}^{\theta} \right) ^{-1}\mathrm{Log}( \hat{\mathbf{R}}_{k}^{\mathrm{T}}\mathbf{R}_k ) \right)}	
\end{equation}
\begin{equation}
	\phi \left( x \right) =\begin{cases}
		1, \ \mathrm{if}\ x\leq 1\\
		0, \ \mathrm{otherwise}\\
	\end{cases}
\end{equation}
where $\mathbf{t}_k$ and $\mathbf{R}_k$ are the ground-truth pose, $\hat{\mathbf{t}}_k$ and $\hat{\mathbf{R}}_{k}$ are the estimated pose,  $\hat{\mathbf{P}}_{k}^{\mathrm{t}}$ and $\hat{\mathbf{P}}_{k}^{\theta}$ are the scaled covariances or the \emph{global protection levels}. For probabilistic methods, the covariance of a point cloud with $n$ points is given by $\mathrm{diag}\left( \sigma ^2,\sigma ^2,\cdots ,\sigma ^2 \right) $. In our experiments, different covariances of point noises ($\sigma^2$) were tuned for probabilistic methods to verify the performance of their protection levels. Since reliable ground-truth rotation of some sequences is not provided, we only evaluated $\mathrm{CR_{rot}}$ on some specific sequences. The CR results on the M2DGR dataset are shown in Table \ref{tab:cr_trans_m2dgr} and Table \ref{tab:cr_rot_m2dgr}, and the results on the NTU VIRAL dataset are shown in Table \ref{tab:cr_trans_ntuviral}. The $\mathrm{CR_{trans}}$ results of the protection levels estimated by our method outperform PALoc by 35.47\% on the M2DGR dataset, and by 110.94\% on the NTU VIRAL dataset. The $\mathrm{CR_{rot}}$ results of the protection levels estimated by our method outperform PALoc as well by 36.58\% on the M2DGR dataset. Although the method from \cite{brossard_2020} yields considerable CR results, its localization accuracy is limited. The $\mathrm{CR_{trans}}$ and $\mathrm{CR_{rot}}$ results on the SubT-MRS dataset are shown in Table \ref{tab:cr_trans_subtmrs} and Table \ref{tab:cr_rot_subtmrs}. These results indicate that, in terms of estimating protection levels and providing online safety assessment, our deterministic method based on the set-membership filter has a significant advantage over the traditional probabilistic methods. It shows that the low performance of the protection levels obtained by the probabilistic methods is not related to the noise parameters but is caused by the inherent characteristics of the Gaussian distribution.

\begin{table*}[htp]
	\caption{$\mathrm{CR}_\mathrm{trans}$ [\%] ($\uparrow$) on the M2DGR dataset.}
	\label{tab:cr_trans_m2dgr}
	\setlength{\tabcolsep}{3.1mm}
	\setlength{\extrarowheight}{0.1mm}
	\centering
	\begin{tabular}{lccccccccccccc}
		\toprule
		& \rotatebox[origin=c]{90}{\shortstack{FAST-LIO2\\($\sigma^2$=0.001)}} & \rotatebox[origin=c]{90}{\shortstack{FAST-LIO2\\($\sigma^2$=0.1)}} & \rotatebox[origin=c]{90}{\shortstack{FAST-LIO2\\($\sigma^2$=1)}} & \rotatebox[origin=c]{90}{\shortstack{LIO-EKF\\($\sigma^2$=0.001)}} & \rotatebox[origin=c]{90}{\shortstack{LIO-EKF\\($\sigma^2$=0.1)}} & \rotatebox[origin=c]{90}{\shortstack{LIO-EKF\\($\sigma^2$=1)}} & \rotatebox[origin=c]{90}{LIO-SAM} & \rotatebox[origin=c]{90}{DLIO} & \rotatebox[origin=c]{90}{Brossard \it{et al.}} & \rotatebox[origin=c]{90}{\shortstack{PALoc\\($\sigma^2$=0.001)}} & \rotatebox[origin=c]{90}{\shortstack{PALoc\\($\sigma^2$=0.1)}} & \rotatebox[origin=c]{90}{\shortstack{PALoc\\($\sigma^2$=1)}} & \rotatebox[origin=c]{90}{Ours} \\
		\midrule
		street\_03 & 0.028 & 0.028 & 0.028 & 0.000 & 0.113 & 1.387 & $-$ & $-$ & 76.516 & 7.932 & 88.725 & 96.997 & \textbf{100} \\
		street\_08 & 0.020 & 0.020 & 0.020 & 0.000 & 0.326 & 1.874 & $-$ & $-$ & 72.183 & 44.834 & 96.331 & 98.757 & \textbf{100} \\
		gate\_01 & 0.000 & 0.000 & 0.000 & 0.000 & 0.000 & 0.756 & $-$ & $-$ & 87.217 & 10.192 & 80.664 & 98.019 & \textbf{100} \\
		gate\_03 & 0.035 & 0.035 & 0.035 & 0.000 & 0.106 & 0.600 & $-$ & $-$ & 87.778 & 63.190 & 95.477 & 99.081 & \textbf{100} \\
		hall\_01 & 0.000 & 0.000 & 0.000 & 0.000 & 1.053 & 4.211 & $-$ & $-$ & 85.263 & 0.000 & 53.684 & 80.000 & \textbf{100} \\
		hall\_02 & 0.000 & 0.000 & 0.000 & 0.000 & 0.000 & 0.000 & $-$ & $-$ & 87.500 & 0.000 & 16.667 & 54.167 & \textbf{100} \\
		hall\_03 & 0.000 & 0.000 & 0.000 & 0.000 & 0.000 & 0.000 & $-$ & $-$ & 80.000 & 0.000 & 12.245 & 8.163 & \textbf{100} \\
		room\_01 & 0.150 & 0.150 & 0.151 & 0.000 & 1.349 & 15.940 & $-$ & $-$ & 89.805 & 0.000 & 19.127 & 71.795 & \textbf{100} \\
		room\_02 & 0.152 & 0.152 & 0.152 & 0.000 & 2.266 & 34.894 & $-$ & $-$ & 88.520 & 1.368 & 41.578 & 89.210 & \textbf{100} \\
		room\_03 & 0.080 & 0.080 & 0.080 & 0.000 & 0.239 & 11.358 & $-$ & $-$ & 97.375 & 0.000 & 7.962 & 41.992 & \textbf{100} \\ \midrule
		Average & 0.047 & 0.047 & 0.047 & 0.000 & 0.545 & 7.102 & $-$ & $-$ & 85.216 & 12.752 & 51.246 & 73.818 & \textbf{100} \\
		\bottomrule
	\end{tabular}
\end{table*}

\begin{table*}[htp]
	\caption{$\mathrm{CR}_\mathrm{rot}$ [\%] ($\uparrow$) on the M2DGR dataset.}
	\label{tab:cr_rot_m2dgr}
	\setlength{\tabcolsep}{3.1mm}
	\setlength{\extrarowheight}{0.1mm}
	\centering
	\begin{tabular}{lccccccccccccc}
		\toprule
		& \rotatebox[origin=c]{90}{\shortstack{FAST-LIO2\\($\sigma^2$=0.001)}} & \rotatebox[origin=c]{90}{\shortstack{FAST-LIO2\\($\sigma^2$=0.1)}} & \rotatebox[origin=c]{90}{\shortstack{FAST-LIO2\\($\sigma^2$=1)}} & \rotatebox[origin=c]{90}{\shortstack{LIO-EKF\\($\sigma^2$=0.001)}} & \rotatebox[origin=c]{90}{\shortstack{LIO-EKF\\($\sigma^2$=0.1)}} & \rotatebox[origin=c]{90}{\shortstack{LIO-EKF\\($\sigma^2$=1)}} & \rotatebox[origin=c]{90}{LIO-SAM} & \rotatebox[origin=c]{90}{DLIO}
		& \rotatebox[origin=c]{90}{Brossard \it{et al.}} & \rotatebox[origin=c]{90}{\shortstack{PALoc\\($\sigma^2$=0.001)}} & \rotatebox[origin=c]{90}{\shortstack{PALoc\\($\sigma^2$=0.1)}} & \rotatebox[origin=c]{90}{\shortstack{PALoc\\($\sigma^2$=1)}} & \rotatebox[origin=c]{90}{Ours} \\
		\midrule
		street\_03 & 1.586 & 5.466 & 40.176 & 0.000 & 0.000 & 0.000 & $-$ & $-$ & 83.938 & 0.368 & 19.773 & 92.606 & \textbf{100} \\
		street\_08 & 0.550 & 24.414 & 76.793 & 0.000 & 0.000 & 0.000 & $-$ & $-$ & 59.344 & 4.239 & 79.698 & 99.796 & \textbf{100} \\
		gate\_01 & 0.524 & 0.524 & 0.641 & 0.000 & 0.000 & 0.000 & $-$ & $-$ & 100 & 0.058 & 0.524 & 0.466 & \textbf{100} \\
		gate\_03 & 1.308 & 24.302 & 50.548 & 0.000 & 0.000 & 0.000 & $-$ & $-$ & 97.916 & 2.511 & 60.848 & 100 & \textbf{100} \\ \midrule
		Average & 0.992 & 13.677 & 42.040 & 0.000 & 0.000 & 0.000 & $-$ & $-$ & 85.300 & 1.794 & 40.211 & 73.217 & \textbf{100} \\
		\bottomrule
	\end{tabular}
\end{table*}

\begin{table*}[htp]
	\centering
	\caption{$\mathrm{CR}_\mathrm{trans}$ [\%] ($\uparrow$) on the NTU VIRAL dataset.}
	\label{tab:cr_trans_ntuviral}
	\setlength{\tabcolsep}{3.1mm}
	\setlength{\extrarowheight}{0.1mm}
	\begin{tabular}{lccccccccccccc}
		\toprule
		& \rotatebox[origin=c]{90}{\shortstack{FAST-LIO2\\($\sigma^2$=0.001)}} & \rotatebox[origin=c]{90}{\shortstack{FAST-LIO2\\($\sigma^2$=0.1)}} & \rotatebox[origin=c]{90}{\shortstack{FAST-LIO2\\($\sigma^2$=1)}} & \rotatebox[origin=c]{90}{\shortstack{LIO-EKF\\($\sigma^2$=0.001)}} & \rotatebox[origin=c]{90}{\shortstack{LIO-EKF\\($\sigma^2$=0.1)}} & \rotatebox[origin=c]{90}{\shortstack{LIO-EKF\\($\sigma^2$=1)}} & \rotatebox[origin=c]{90}{LIO-SAM} & \rotatebox[origin=c]{90}{DLIO}
		& \rotatebox[origin=c]{90}{Brossard \it{et al.}} & \rotatebox[origin=c]{90}{\shortstack{PALoc\\($\sigma^2$=0.001)}} & \rotatebox[origin=c]{90}{\shortstack{PALoc\\($\sigma^2$=0.1)}} & \rotatebox[origin=c]{90}{\shortstack{PALoc\\($\sigma^2$=1)}} & \rotatebox[origin=c]{90}{Ours} \\
		\midrule
		eee\_01 & 0.000 & 0.000 & 0.000 & 0.000 & 3.607 & 24.136 & $-$ & $-$ & 84.560 & 68.930 & 44.441 & 84.105 & \textbf{100} \\
		eee\_02 & 0.000 & 0.000 & 0.000 & 0.036 & 4.608 & 33.801 & $-$ & $-$ & 72.030 & 47.084 & 8.675 & 43.629 & \textbf{100} \\
		eee\_03 & 0.000 & 0.000 & 0.000 & 0.000 & 1.132 & 14.048 & $-$ & $-$ & 53.298 & 38.774 & 30.646 & 89.540 & \textbf{100} \\
		nya\_01 & 0.000 & 0.000 & 0.000 & 0.000 & 2.004 & 40.827 & $-$ & $-$ & 68.809 & 27.123 & 20.638 & 34.894 & \textbf{100} \\
		nya\_02 & 0.000 & 0.000 & 0.000 & 0.000 & 3.137 & 0.130 & $-$ & $-$ & 61.566 & 6.743 & 66.209 & 69.579 & \textbf{100} \\
		nya\_03 & 0.000 & 0.000 & 0.000 & 0.000 & 3.180 & 24.224 & $-$ & $-$ & 53.452 & 1.319 & 24.722 & 16.266 & \textbf{100} \\
		sbs\_01 & 0.000 & 0.000 & 0.000 & 0.000 & 2.776 & 32.811 & $-$ & $-$ & 84.358 & 71.632 & 29.826 & 88.020 & \textbf{100} \\
		sbs\_02 & 0.000 & 0.000 & 0.000 & 0.000 & 2.162 & 8.549 & $-$ & $-$ & 71.368 & 60.995 & 0.000 & 0.000 & \textbf{100} \\
		sbs\_03 & 0.000 & 0.000 & 0.000 & 0.000 & 3.241 & 33.724 & $-$ & $-$ & 72.239 & 60.956 & 0.000 & 0.619 & \textbf{100} \\ \midrule
		Average & 0.000 & 0.000 & 0.000 & 0.004 & 2.872 & 23.583 & $-$ & $-$ & 69.076 & 42.617 & 25.017 & 47.406 & \textbf{100} \\
		\bottomrule
	\end{tabular}
\end{table*}

To further demonstrate the superiority of the protection levels obtained by our system, we decompose the estimated protection levels along three axes, and the results are depicted in Fig. \ref{fig:bound_m2dgr}. It turns out that the protection levels obtained by PALoc cannot effectively cover the ground-truth trajectories, thus making it impossible to conduct an effective online assessment of the safety and reliability of the localization system. The results also show that the protection levels estimated by our system can effectively encompass the localization errors. It indicates that when the ground truth is unknown in real-world applications, our deterministic protection level can be used as a bound on the localization error under the assumed noise bounds, thereby providing a safety reference for downstream tasks.

\begin{figure}[tp]
	\centering
	\includegraphics[width=3.2in]{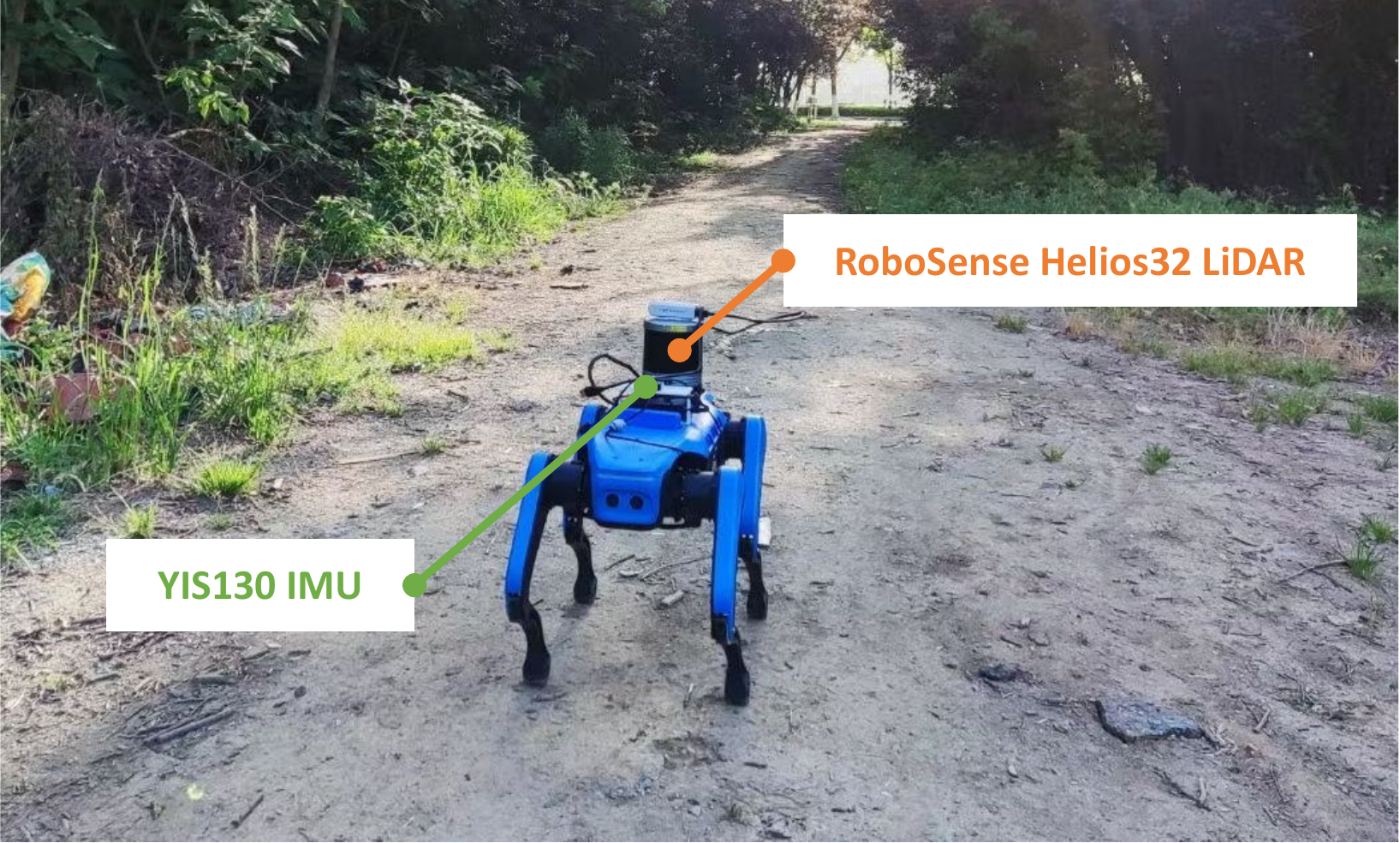}
	\caption{Our quadruped robot platform. The RoboSense Helios32 LiDAR and the YIS130 IMU were used as the sensors. The quadruped robot was remotely controlled and walked along multiple different trajectories in the campus environment.}
	\label{fig:quadruped_robot}
\end{figure}

\begin{figure}[tp]
	\centering
	\includegraphics[width=3.2in]{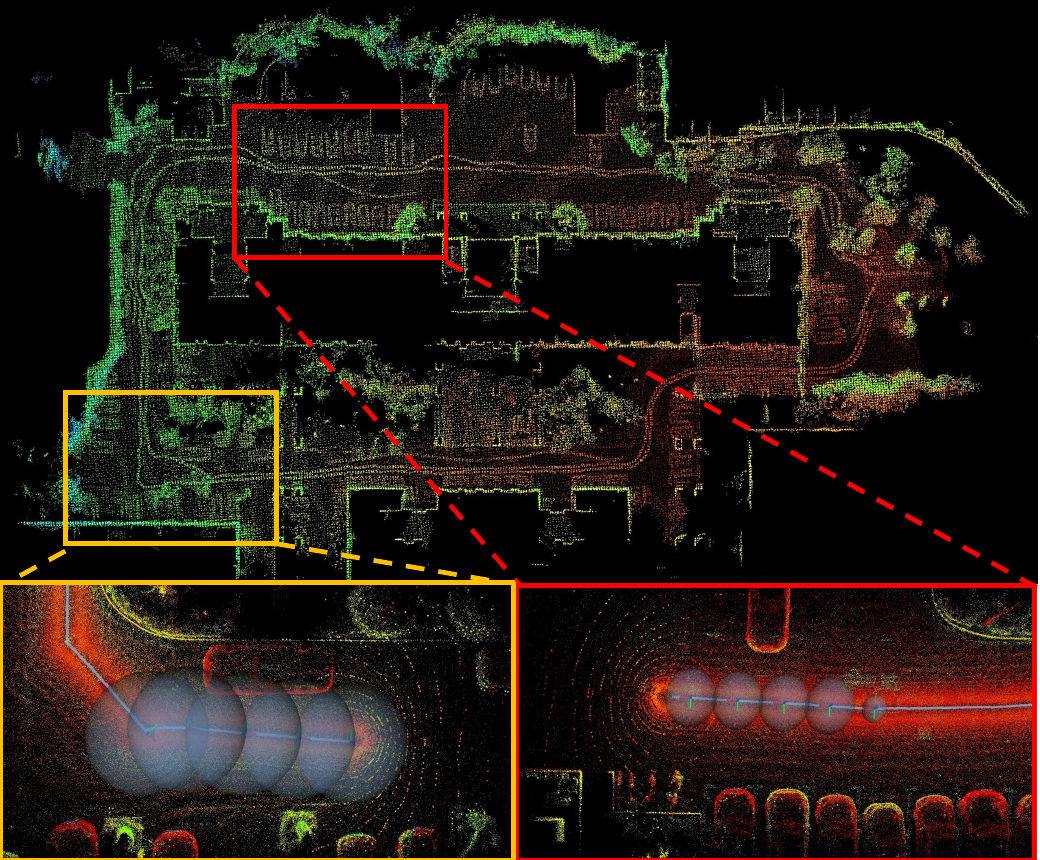}
	\caption{Estimated \emph{global protection levels} and mapping result of the \textit{building} sequence. Our system not only provides location estimations and deterministic protection levels, but also yields fine-grained maps.}
	\label{fig:liwenzheng}
\end{figure}

\begin{table*}[htp]
	\centering
	\caption{$\mathrm{CR}_\mathrm{trans}$ [\%] ($\uparrow$) on the SubT-MRS dataset.}
	\label{tab:cr_trans_subtmrs}
	\setlength{\tabcolsep}{2.4mm}
	\setlength{\extrarowheight}{0.6mm}
	\begin{tabular}{lcccccccccccccc}
		\toprule
		& \rotatebox[origin=c]{90}{\shortstack{FAST-LIO2\\($\sigma^2$=0.001)}} & \rotatebox[origin=c]{90}{\shortstack{FAST-LIO2\\($\sigma^2$=0.1)}} & \rotatebox[origin=c]{90}{\shortstack{FAST-LIO2\\($\sigma^2$=1)}} & \rotatebox[origin=c]{90}{\shortstack{LIO-EKF\\($\sigma^2$=0.001)}} & \rotatebox[origin=c]{90}{\shortstack{LIO-EKF\\($\sigma^2$=0.1)}} & \rotatebox[origin=c]{90}{\shortstack{LIO-EKF\\($\sigma^2$=1)}} & \rotatebox[origin=c]{90}{LIO-SAM} & \rotatebox[origin=c]{90}{DLIO}
		& \rotatebox[origin=c]{90}{Brossard \it{et al.}} & \rotatebox[origin=c]{90}{\shortstack{PALoc\\($\sigma^2$=0.001)}} & \rotatebox[origin=c]{90}{\shortstack{PALoc\\($\sigma^2$=0.1)}} & \rotatebox[origin=c]{90}{\shortstack{PALoc\\($\sigma^2$=1)}} & \rotatebox[origin=c]{90}{\shortstack{Ours\\w/o deg$^1$}} & \rotatebox[origin=c]{90}{Ours} \\
		\midrule
		Urban\_UGV1 & 0.000 & 0.000 & 0.000 & 0.000 & 47.518 & 1.793 & $-$ & $-$ & 73.385 & 56.068 & 50.193 & 27.034 & 100 & \textbf{100} \\
		Urban\_UGV2 & 0.000 & 0.000 & 0.000 & 0.000 & 73.158 & 48.933 & $-$ & $-$ & 79.710 & 0.000 & 7.879 & 7.696 & 99.089 & \textbf{100} \\
		Laurel\_Caverns & 0.000 & 0.000 & 0.000 & 0.000 & 20.395 & 5.403 & $-$ & $-$ & 72.055 & 48.948 & 65.724 & 68.379 & 0.486 & \textbf{100} \\
		Long\_Corridor & 0.000 & 0.000 & 0.000 & 0.000 & 52.240 & 0.903 & $-$ & $-$ & 71.821 & 43.461 & 45.592 & 45.376 & 0.000 & \textbf{100} \\
		Multi\_Floor & 0.000 & 0.000 & 0.000 & 0.000 & 29.625 & 0.536 & $-$ & $-$ & 87.366 & 45.935 & 25.880 & 44.523 & 56.889 & \textbf{100} \\ \midrule
		Average & 0.000 & 0.000 & 0.000 & 0.000 & 44.587 & 11.514 & $-$ & $-$ & 76.867 & 38.882 & 39.054 & 38.602 & 51.293 & \textbf{100} \\
		\bottomrule
	\end{tabular}
	\begin{tablenotes}[flushleft]
		\footnotesize
		\item $^1$ Without degeneracy management.
	\end{tablenotes}
\end{table*}

\begin{table*}[htp]
	\centering
	\caption{$\mathrm{CR}_\mathrm{rot}$ [\%] ($\uparrow$) on the SubT-MRS dataset.}
	\label{tab:cr_rot_subtmrs}
	\setlength{\tabcolsep}{2.4mm}
	\setlength{\extrarowheight}{0.6mm}
	\begin{tabular}{lcccccccccccccc}
		\toprule
		& \rotatebox[origin=c]{90}{\shortstack{FAST-LIO2\\($\sigma^2$=0.001)}} & \rotatebox[origin=c]{90}{\shortstack{FAST-LIO2\\($\sigma^2$=0.1)}} & \rotatebox[origin=c]{90}{\shortstack{FAST-LIO2\\($\sigma^2$=1)}} & \rotatebox[origin=c]{90}{\shortstack{LIO-EKF\\($\sigma^2$=0.001)}} & \rotatebox[origin=c]{90}{\shortstack{LIO-EKF\\($\sigma^2$=0.1)}} & \rotatebox[origin=c]{90}{\shortstack{LIO-EKF\\($\sigma^2$=1)}} & \rotatebox[origin=c]{90}{LIO-SAM} & \rotatebox[origin=c]{90}{DLIO}
		& \rotatebox[origin=c]{90}{Brossard \it{et al.}} & \rotatebox[origin=c]{90}{\shortstack{PALoc\\($\sigma^2$=0.001)}} & \rotatebox[origin=c]{90}{\shortstack{PALoc\\($\sigma^2$=0.1)}} & \rotatebox[origin=c]{90}{\shortstack{PALoc\\($\sigma^2$=1)}} & \rotatebox[origin=c]{90}{\shortstack{Ours\\w/o deg$^1$}} & \rotatebox[origin=c]{90}{Ours} \\
		\midrule
		Urban\_UGV1 & 0.020 & 0.256 & 14.716 & 0.000 & 0.000 & 0.000 & $-$ & $-$ & 99.862 & 10.540 & 89.055 & 66.773 & {100} & \textbf{100} \\
		Urban\_UGV2 & 0.003 & 0.054 & 1.270 & 0.000 & 0.000 & 0.000 & $-$ & $-$ & 99.994 & 3.853 & 23.437 & 48.422 & {100} & \textbf{100} \\
		Laurel\_Caverns & 70.484 & 79.607 & 89.508 & 0.000 & 0.000 & 0.000 & $-$ & $-$ & 97.082 & 56.650 & 85.836 & 97.419 & {100} & \textbf{100} \\
		Long\_Corridor & 0.036 & 20.701 & 44.689 & 0.000 & 0.000 & 0.000 & $-$ & $-$ & 98.627 & 51.048 & 45.665 & 55.383 & 84.821 & \textbf{100} \\
		Multi\_Floor & 0.024 & 8.909 & 46.981 & 0.000 & 0.000 & 0.000 & $-$ & $-$ & 99.537 & 49.367 & 48.455 & 100 & 100 & \textbf{100} \\ \midrule
		Average & 14.113 & 21.905 & 39.433 & 0.000 & 0.000 & 0.000 & $-$ & $-$ & 99.020 & 34.292 & 58.490 & 73.599 & 96.964 & \textbf{100} \\
		\bottomrule
	\end{tabular}
	\begin{tablenotes}[flushleft]
		\footnotesize
		\item $^1$ Without degeneracy management.
	\end{tablenotes}
\end{table*}

\subsection{Real-World Results}
In addition to the public datasets, to verify the effectiveness of the protection levels estimated by our system, experiments using a quadruped robot equipped with the RoboSense Helios32 LiDAR and the YIS130 IMU were conducted. The robot platform is shown in Fig. \ref{fig:quadruped_robot}. 

\begin{figure}[tp]
	\centering
	\includegraphics[width=3in]{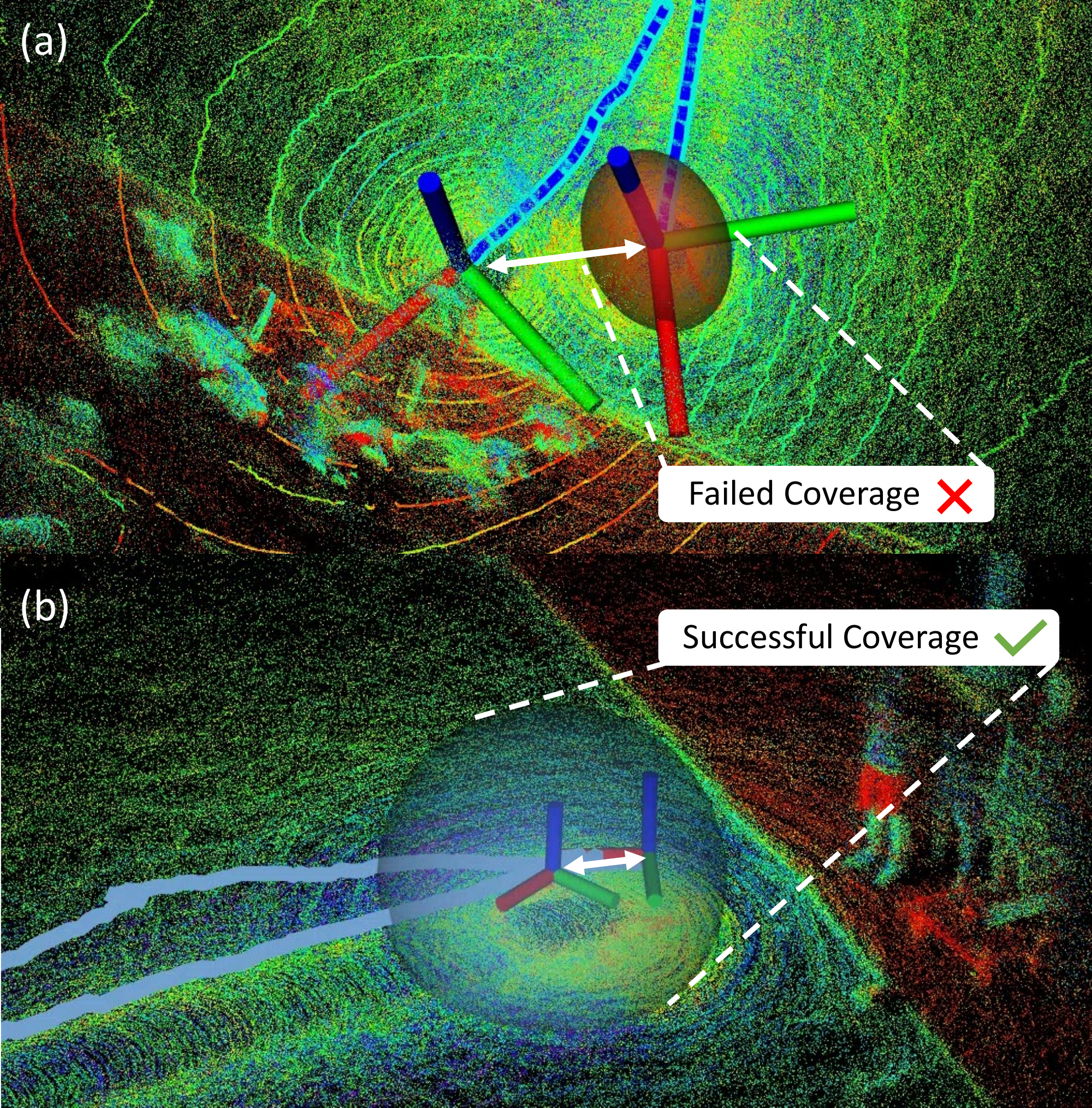}
	\caption{Ending protection levels of the \textit{lawn} sequence from different methods. The results show that when the robot returned to the origin, the protection level obtained by our system can effectively cover the errors, while the one obtained by PALoc cannot. (a) PALoc ($\sigma ^2=0.001$). (b) Ours.}
	\label{fig:end_coverage}
\end{figure}

The estimated protection levels and the mapping result of the \textit{building} sequence are shown in Fig. \ref{fig:liwenzheng}. In the real-world experiments using the quadruped robot, we use the end-to-end error to evaluate the accuracy of the localization systems. To fairly verify the end-to-end accuracy of the LIO methods, all loop closure modules were turned off. The end-to-end translational errors are shown in Table \ref{tab:end_to_end_error}. The abilities to cover the ending translational errors are shown in Table \ref{tab:end_to_end_coverage} and Fig. \ref{fig:end_coverage}. The results show that even without optimal estimation results, the accuracy of our system is still improved compared with some SOTA LIOs. More importantly, when the robot moved for a long time and returned to the origin, the protection level obtained by our system can effectively express the uncertainty of the final localization result. This indicates that our deterministic protection level can effectively be utilized to ensure the safety of the autonomous operation of robots.

\begin{table}[htp]
	\caption{End-to-end translational error [m] ($\downarrow$) in different sequences with quadruped robot.}
	\label{tab:end_to_end_error}
	\setlength{\tabcolsep}{4mm}
	\setlength{\extrarowheight}{0.6mm}
	\centering
	\begin{tabular}{lccc}
		\toprule
		Methods         & \textit{lawn}  & \textit{building} & \textit{sidewalk} \\ \midrule
		FAST-LIO2 ($\sigma ^2=0.001$) & 1.004          & 2.177             & 0.893             \\
		FAST-LIO2 ($\sigma ^2=0.1$) & \textbf{0.714}          & $\times$             & 2.339             \\
		FAST-LIO2 ($\sigma ^2=1$) & $\times$          & $\times$             & $\times$             \\
		LIO-EKF         & 1.840          & 3.730             & 2.168             \\ \midrule
		LIO-SAM         & 0.989          & 3.601             & $\times$          \\
		DLIO & $\times$ & \textbf{0.661} &  4.442
		\\
		Brossard \it{et al.} & 18.919 & 35.408 &  21.825 \\
		PALoc ($\sigma ^2=0.001$)     & 2.270          & {1.177}    & 0.929             \\
		PALoc ($\sigma ^2=0.1$)     & $\times$          & $\times$    & \textbf{0.708}             \\
		PALoc ($\sigma ^2=1$)     & $\times$          & $\times$    & $\times$             \\ \midrule
		Ours            & {0.953} & {1.496}             & {0.710}    \\ \bottomrule
	\end{tabular}
\end{table}

\begin{table}[htp]
	\caption{Ending translational coverage in different sequences with quadruped robot.}
	\label{tab:end_to_end_coverage}
	\setlength{\tabcolsep}{4mm}
	\setlength{\extrarowheight}{0.6mm}
	\centering
	\begin{tabular}{lccc}
		\toprule
		Methods         & \textit{lawn}  & \textit{building} & \textit{sidewalk} \\ \midrule
		FAST-LIO2 ($\sigma ^2=0.001$) & $\times$          & $\times$             & $\times$             \\
		FAST-LIO2 ($\sigma ^2=0.1$) & $\times$          & $\times$             & $\times$             \\
		FAST-LIO2 ($\sigma ^2=1$) & $\times$          & $\times$             & $\times$            \\
		LIO-EKF & $\times$          & $\times$             & $\times$             \\ \midrule
		LIO-SAM & $-$          & $-$             & $-$             \\
		DLIO & $-$ & $-$ & $-$
		\\
		Brossard \it{et al.} & $\times$ & $\checkmark$ &  $\times$ \\
		PALoc ($\sigma ^2=0.001$)     & $\times$          & $\checkmark$    & $\times$             \\
		PALoc ($\sigma ^2=0.1$)     & $\times$          & $\times$    & $\checkmark$             \\
		PALoc ($\sigma ^2=1$)     & $\times$          & $\times$    & $\times$             \\ \midrule
		Ours            & $\checkmark$ & $\checkmark$             & $\checkmark$    \\ \bottomrule
	\end{tabular}
\end{table}

\begin{figure}[tp]
	\centering
	\includegraphics[width=2in]{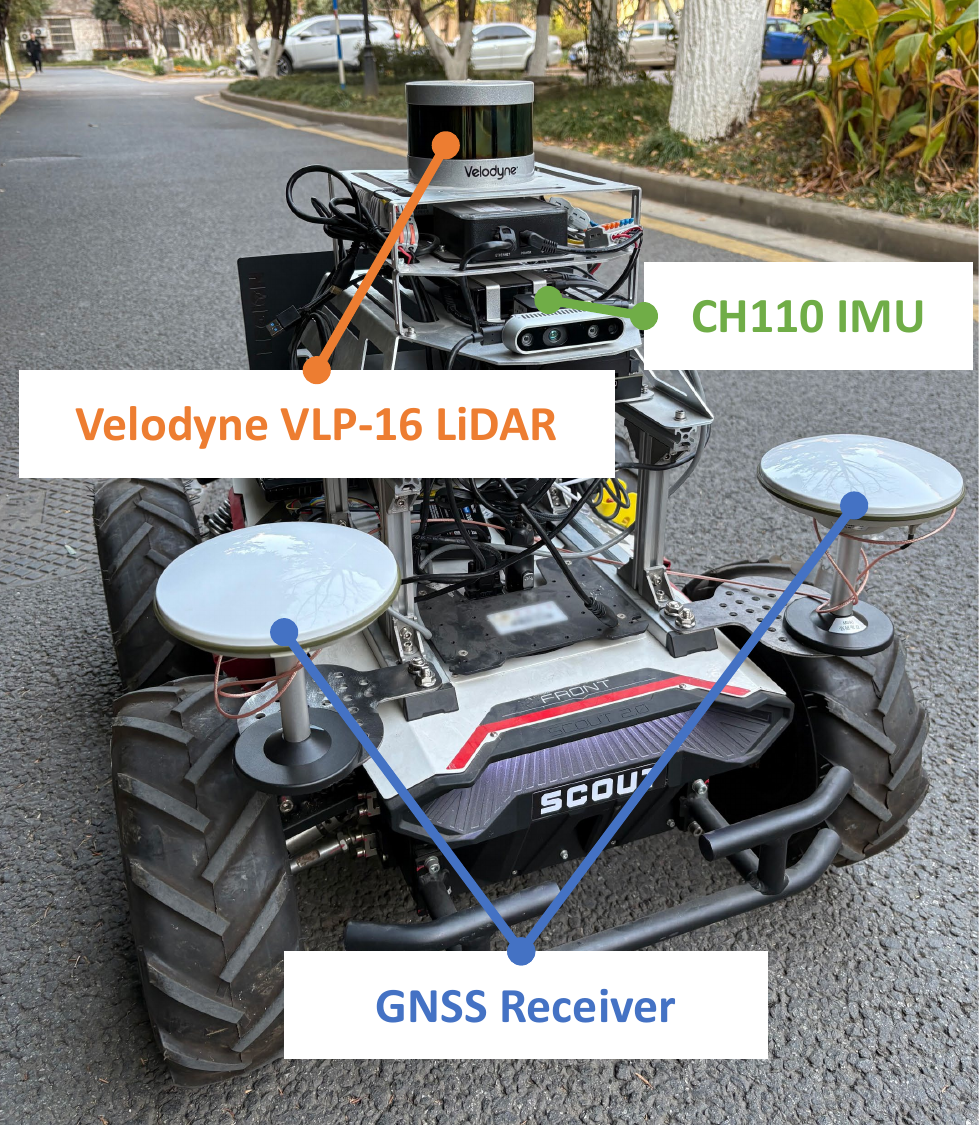}
	\caption{Our wheeled mobile robot platform. The Velodyne VLP-16 LiDAR and CH110 IMU were used as the sensors. The real-time kinematic GNSS was equipped to provide ground truth.}
	\label{fig:wheeled_robot}
\end{figure}

To verify the cross-platform capability of our system, as shown in Fig. \ref{fig:wheeled_robot}, we also conducted experiments using a wheeled mobile robot equipped with Velodyne VLP-16 LiDAR and CH110 IMU. The real-time kinematic GNSS was used to provide ground truth for the robot. The ATE results and the CR results are shown in Table \ref{tab:ate2} and Table \ref{tab:coverage2}. The results show that, even though the platforms are different, our system still performs exceptionally well. In the experiments conducted on the wheeled mobile robot platform, the localization accuracy our system achieved was close to the SOTA probabilistic methods. More importantly, the deterministic protection levels provided by our system can accurately cover the ground truth.

\begin{table}[htp]
	\caption{ATE [m] ($\downarrow$) in different sequences with wheeled mobile robot.}
	\label{tab:ate2}
	\setlength{\tabcolsep}{4mm}
	\setlength{\extrarowheight}{0.6mm}
	\centering
	\begin{tabular}{lccc}
		\toprule
		Methods         & \textit{yard}  & \textit{garden} & \textit{fountain} \\ \midrule
		FAST-LIO2 ($\sigma ^2=0.001$) & 0.344          & 0.658 & 0.828          \\
		FAST-LIO2 ($\sigma ^2=0.1$) & $\times$          & 0.731 & $\times$             \\
		FAST-LIO2 ($\sigma ^2=1$) & $\times$          & $\times$             & $\times$             \\
		LIO-EKF         & 0.311 & 0.660 & 0.841 \\ \midrule
		LIO-SAM         & {0.215} & \textbf{0.567} & \textbf{0.773} \\
		DLIO & \textbf{0.118} & 0.576 &  0.807
		\\
		Brossard \it{et al.} & 5.352 & 8.130 &  8.650 \\
		PALoc ($\sigma ^2=0.001$)     & 0.325 & 0.665 & 0.908 \\
		PALoc ($\sigma ^2=0.1$)     & $\times$          & $\times$    & $\times$             \\
		PALoc ($\sigma ^2=1$)     & $\times$          & $\times$    & $\times$             \\ \midrule
		Ours            & {0.129} & {0.635}             &{0.828}    \\ \bottomrule
	\end{tabular}
\end{table}

\begin{table}[htp]
	\centering
	\caption{$\mathrm{CR}_\mathrm{trans}$ [\%] ($\uparrow$) in different sequences with wheeled mobile robot.}
	\label{tab:coverage2}
	\setlength{\tabcolsep}{4mm}
	\setlength{\extrarowheight}{0.6mm}
	\centering
	\begin{tabular}{lccc}
		\toprule
		Methods         & \textit{yard}  & \textit{garden} & \textit{fountain} \\ \midrule
		FAST-LIO2 ($\sigma ^2=0.001$) & 0.000          & 0.000             & 0.000             \\
		FAST-LIO2 ($\sigma ^2=0.1$) & $\times$          & 0.000             & $\times$             \\
		FAST-LIO2 ($\sigma ^2=1$) & $\times$          & $\times$             & $\times$            \\
		LIO-EKF & 0.000          & 0.000             & 0.000             \\ \midrule
		LIO-SAM & $-$          & $-$             & $-$             \\
		DLIO & $-$ & $-$ & $-$
		\\
		Brossard \it{et al.} & 40.041 & 74.473 &  62.784 \\
		PALoc ($\sigma ^2=0.001$)     & 0.824          & 0.000    & 0.000             \\
		PALoc ($\sigma ^2=0.1$)     & $\times$          & $\times$    & $\times$             \\
		PALoc ($\sigma ^2=1$)     & $\times$          & $\times$    & $\times$             \\ \midrule
		Ours            & \textbf{100} & \textbf{100}             & \textbf{100}    \\ \bottomrule
	\end{tabular}
\end{table}

%\begin{figure*}[t]
%	\centering
%	\subfloat[\textit{yard}]
%	{\includegraphics[width=2.2in]{pics/results/yard}}
%	\hspace{2px}
%	\subfloat[\textit{garden}]
%	{\includegraphics[width=2.2in]{pics/results/garden}}
%	\hspace{2px}
%	\subfloat[\textit{fountain}]
%	{\includegraphics[width=2.2in]{pics/results/fountain}}
%	\caption{Estimated protection levels and trajectories from our system and PALoc ($\sigma ^2=0.001$), and the ground truth using different sequences with our wheeled mobile robot. The protection levels obtained by our system can cover the ground truth necessarily. However, even though the three-sigma rule represents the uncertainty with an extremely high probability, there are still many ground truth that fall outside the protection levels.}
%	\label{fig:bound_scout}
%\end{figure*}

Finally, to verify the adaptability of our system to different types of LiDARs and environments, we conducted indoor experiments using the solid-state Livox Mid-360 and its built-in IMU. The optical motion capture system was used to provide the ground truth. The ATE results are shown in Table \ref{tab:ate3}. The results show that in the indoor environment, the localization accuracy results of various methods are comparable. The CR results are listed in Table \ref{tab:coverage3}. Since both PALoc and our system can achieve 100\% CR, we use the average interval length (AIL) \cite{stutts_2023} to measure the conservatism of the protection levels. 
%For our deterministic protection levels, the average interval length can be defined as:
%\begin{equation}
%	\mathrm{AIL}_\mathrm{trans}=\frac{1}{K}\sum_{k=1}^K{\frac{1}{3}\sum_{i=1}^3{2\sqrt{\hat{\mathbf{P}}_{k}^{\mathrm{t},\mathrm{glb}}\left( i,i \right)}}}
%\end{equation}
%\begin{equation}
%	\mathrm{AIL}_\mathrm{rot}=\frac{1}{K}\sum_{k=1}^K{\frac{1}{3}\sum_{i=1}^3{2\sqrt{\hat{\mathbf{P}}_{k}^{\theta,\mathrm{glb}}\left( i,i \right)}}}
%\end{equation}where $\hat{\mathbf{P}}_{k}^{\mathrm{t},\mathrm{glb}}$ and  $\hat{\mathbf{P}}_{k}^{\theta,\mathrm{glb}}$ are the shape matrices, $\sqrt{\hat{\mathbf{P}}_{k}^{\mathrm{t},\mathrm{glb}}( i,i )}$ and $\sqrt{\hat{\mathbf{P}}_{k}^{\theta,\mathrm{glb}}( i,i )}$ are the radii of the intervals \cite{scholte_2003}. For other probabilistic protection levels, the average interval length can be defined based on the three-sigma rule:
%\begin{equation}
%	\mathrm{AIL}_\mathrm{trans}=\frac{1}{K}\sum_{k=1}^K{\frac{1}{3}\sum_{i=1}^3{2\times3\sqrt{\hat{\mathbf{P}}_{k}^{\mathrm{t}}\left( i,i \right)}}}
%\end{equation}
%\begin{equation}
%	\mathrm{AIL}_\mathrm{rot}=\frac{1}{K}\sum_{k=1}^K{\frac{1}{3}\sum_{i=1}^3{2\times3\sqrt{\hat{\mathbf{P}}_{k}^{\theta}\left( i,i \right)}}}
%\end{equation}where $\hat{\mathbf{P}}_{k}^{\mathrm{t}}$ and $\hat{\mathbf{P}}_{k}^{\theta}$ are the estimated covariances of the translation and rotation, $\sqrt{\hat{\mathbf{P}}_{k}^{\mathrm{t}}( i,i )}$ and  $\sqrt{\hat{\mathbf{P}}_{k}^{\theta}( i,i )}$ are the standard deviations. 
In practical applications, the conservatism of the protection level plays a decisive role in the performance of the autonomous navigation system. The protection levels should correctly cover localization errors with a small degree of conservatism. Table \ref{tab:coverage3} indicates that although both PALoc and our system achieved 100\% CR, the protection level obtained by our system is less conservative. Compared with PALoc, our system reduced the conservatism by 47.38\%. In practical applications, this small but precise protection level can provide better safety references for mobile robots.

\begin{table}[tp]
	\centering
	\caption{ATE [m] ($\downarrow$) in different sequences with Livox Mid-360.}
	\label{tab:ate3}
	\setlength{\tabcolsep}{4mm}
	\setlength{\extrarowheight}{0.6mm}
	\centering
	\begin{tabular}{lccc}
		\toprule
		Methods         & \textit{seq. 1}  & \textit{seq. 2} & \textit{seq. 3} \\ \midrule
		FAST-LIO2 ($\sigma ^2=0.001$) & 0.336          & 0.255 & 0.223          \\
		FAST-LIO2 ($\sigma ^2=0.1$) & 0.333          & 0.253 & 0.222             \\
		FAST-LIO2 ($\sigma ^2=1$) & 0.331          & {0.251}             & {0.220}             \\
		LIO-EKF         & $\times$ & $\times$ & $\times$ \\ \midrule
		LIO-SAM         & $-^1$ & $-^1$ & $-^1$ \\
		DLIO & 0.330 & \textbf{0.247} &  \textbf{0.181}
		\\ 
		Brossard \it{et al.} & 0.355 & 0.458 &  0.276 \\
		PALoc ($\sigma ^2=0.001$)     & 0.336 & 0.255 & 0.223 \\
		PALoc ($\sigma ^2=0.1$)     & 0.332          & 0.255    & 0.223             \\
		PALoc ($\sigma ^2=1$)     & \textbf{0.255}          & {0.251}    & {0.220}             \\ \midrule
		Ours            & {0.351} & {0.273}             &{0.230}    \\ \bottomrule
	\end{tabular}
	\begin{tablenotes}[]
		\footnotesize
		\item $^1$ Not suitable for the solid-state Livox Mid-360.
	\end{tablenotes}
\end{table}

\begin{table}[htp]
	\centering
	\caption{CR [\%] ($\uparrow$) / AIL [m] ($\downarrow$) in different sequences with Livox Mid-360.}
	\label{tab:coverage3}
	\setlength{\tabcolsep}{0.4mm}
	\setlength{\extrarowheight}{0.1mm}
	\begin{tabular}{lccccc}
		\toprule
		Methods         & \textit{seq. 1}  & \textit{seq. 2} & \textit{seq. 3} & Cover GT & Avg. AIL \\ \midrule
		\begin{tabular}[c]{@{}l@{}}FAST-LIO2\\($\sigma ^2=0.001$)\end{tabular} & 0.000 / $\times^1$          & 0.000 / $\times^1$             & 0.000 / $\times^1$ & $\times$ & $\times^1$           \\
		\begin{tabular}[c]{@{}l@{}}FAST-LIO2\\($\sigma ^2=0.1$)\end{tabular} & 0.000 / $\times^1$          & 2.435 / $\times^1$             & 0.997 / $\times^1$& $\times$ & $\times^1$             \\
		\begin{tabular}[c]{@{}l@{}}FAST-LIO2\\($\sigma ^2=1$)\end{tabular} & 5.910 / $\times^1$          & 17.217 / $\times^1$             & 14.286 / $\times^1$& $\times$ & $\times^1$            \\
		LIO-EKF & $\times$ / $\times$          & $\times$ / $\times$             & $\times$ / $\times$ & $\times$  & $\times$            \\ \midrule
		LIO-SAM & $\times^2$ / $\times^2$          & $\times^2$ / $\times^2$             & $\times^2$ / $\times^2$& $\times^2$  & $\times^2$             \\
		DLIO & $-$ / $-$ & $-$ / $-$ & $-$ / $-$ & $-$ & $-$ \\
		\begin{tabular}[c]{@{}l@{}}PALoc\\($\sigma ^2=0.001$)\end{tabular} & 74.468 / $\times^1$        & 82.609 / $\times^1$    & 72.425 / $\times^1$& $\times$  & $\times^1$           \\
		\begin{tabular}[c]{@{}l@{}}PALoc\\($\sigma ^2=0.1$)\end{tabular} & \textbf{100} / 3.537          & \textbf{100} / 3.909    & \textbf{100} / 2.729 & $\checkmark$  & 3.392           \\
		\begin{tabular}[c]{@{}l@{}}PALoc\\($\sigma ^2=1$)\end{tabular} & \textbf{100} / 7.954          & \textbf{100} / 8.650    & \textbf{100} / 6.425 & $\checkmark$  & 7.676           \\ \midrule
		Ours            & \textbf{100} / \textbf{1.797} & \textbf{100} / \textbf{1.788}             & \textbf{100} / \textbf{1.770} & $\checkmark$  & \textbf{1.785}   \\ \bottomrule
	\end{tabular}
	\begin{tablenotes}[]
		\footnotesize
		\item $^1$ Since the ground truth cannot be 100\% covered, AIL has no meaning.
		\item $^2$ Not suitable for the solid-state Livox Mid-360.
	\end{tablenotes}
\end{table}

\section{Conclusion}
In this paper, we present a safety-critical LiDAR-inertial odometry with on-manifold deterministic protection levels. To measure the uncertainty of the estimated pose from the point-to-plane ICP algorithm, we derive a closed-form expression between the noise of the point cloud and the uncertainty of the pose estimation via the implicit function theorem and the UBB assumption. Besides, an on-manifold set-membership filter is designed to fuse the IMU measurements and the ICP results for more accurate localization with more robust protection levels. Eventually, the experimental results demonstrate that the protection levels estimated by our system can effectively express the errors between estimated locations and unknown ground truth in safety-critical scenarios, and can provide safety references for downstream autonomous tasks.

\appendices

\section{Derivation of the Point Model With UBB Noise}
\label{sec:appd1}

Combining \eqref{eq:point_R3} and \eqref{eq:point_sep} yields
\begin{equation}
	\begin{aligned}
 		^{L}\mathbf{p}_i=d_i\boldsymbol{\phi }_i&=\left( \tilde{d}_i+n_{i}^{\left( d \right)} \right) \left( \tilde{\boldsymbol{\phi}}_i\boxplus _{\mathbb{S} ^2}\boldsymbol{n}_{i}^{\left( \phi \right)} \right) 
 		\\
 		&=\left( \tilde{d}_i+n_{i}^{\left( d \right)} \right) \left( \mathrm{Exp}\left( \mathbf{N}\left( \tilde{\boldsymbol{\phi}}_i \right) \boldsymbol{n}_{i}^{\left( \phi \right)} \right) \tilde{\boldsymbol{\phi}}_i \right) 
	\end{aligned}
\end{equation}
With $\mathrm{Exp}\left( \cdot \right) \approx \mathbf{I}+\left( \cdot \right) ^{\land}$, one can obtain
\begin{equation}
	\begin{aligned}
		^L\mathbf{p}_i&=\left( \tilde{d}_i+n_{i}^{\left( d \right)} \right) \left( \tilde{\boldsymbol{\phi}}_i+\left( \mathbf{N}\left( \tilde{\boldsymbol{\phi}}_i \right) \boldsymbol{n}_{i}^{\left( \phi \right)} \right) ^{\land}\tilde{\boldsymbol{\phi}}_i \right) +\mathbf{r}_{\exp}^{\mathrm{nl}}
		\\
		&=\tilde{d}_i\tilde{\boldsymbol{\phi}}_i+n_{i}^{\left( d \right)}\tilde{\boldsymbol{\phi}}_i+\tilde{d}_i\left( \mathbf{N}\left( \tilde{\boldsymbol{\phi}}_i \right) \boldsymbol{n}_{i}^{\left( \phi \right)} \right) ^{\land}\tilde{\boldsymbol{\phi}}_i
		\\ 
		& \ \ \ \ +\underset{\mathbf{r}_{\mathrm{p}}^{\mathrm{nl}}}{\underbrace{\underset{\approx 0}{\underbrace{n_{i}^{\left( d \right)}\left( \mathbf{N}\left( \tilde{\boldsymbol{\phi}}_i \right) \boldsymbol{n}_{i}^{\left( \phi \right)} \right) ^{\land}\tilde{\boldsymbol{\phi}}_i}}+\mathbf{r}_{\exp}^{\mathrm{nl}}}}
	\end{aligned}
\end{equation}
where $\mathbf{r}_{\exp}^{\mathrm{nl}}$ is the remainder of the exponential function. Different from the original version given in \cite{yuan_2021}, the remainder here is not omitted. Finally, we have
\begin{equation}
	\begin{aligned}
		^L\mathbf{p}_i&=\tilde{d}_i\tilde{\boldsymbol{\phi}}_i+n_{i}^{\left( d \right)}\tilde{\boldsymbol{\phi}}_i+\tilde{d}_i\left( \mathbf{N}\left( \tilde{\boldsymbol{\phi}}_i \right) \boldsymbol{n}_{i}^{\left( \phi \right)} \right) ^{\land}\tilde{\boldsymbol{\phi}}_i+\mathbf{r}_{\mathrm{p}}^{\mathrm{nl}}
		\\
		&=\tilde{d}_i\tilde{\boldsymbol{\phi}}_i+n_{i}^{\left( d \right)}\tilde{\boldsymbol{\phi}}_i-\tilde{d}_i\tilde{\boldsymbol{\phi}}_{i}^{\land}\mathbf{N}\left( \tilde{\boldsymbol{\phi}}_i \right) \boldsymbol{n}_{i}^{\left( \phi \right)}+\mathbf{r}_{\mathrm{p}}^{\mathrm{nl}}
		\\
		&=\tilde{d}_i\tilde{\boldsymbol{\phi}}_i+\left[ \begin{matrix}
			\tilde{\boldsymbol{\phi}}_i&		-\tilde{d}_i\tilde{\boldsymbol{\phi}}_{i}^{\land}\mathbf{N}\left( \tilde{\boldsymbol{\phi}}_i \right)\\
		\end{matrix} \right] \left[ \begin{array}{c}
			n_{i}^{\left( d \right)}\\
			\boldsymbol{n}_{i}^{\left( \phi \right)}\\
		\end{array} \right] +\mathbf{r}_{\mathrm{p}}^{\mathrm{nl}}
	\end{aligned}
\end{equation}

\section{Derivation of the Jacobian Matrix in the ICP Uncertainty Resolving}
\label{sec:appd2}
Denote the objective function in \eqref{eq:last_optimization} as $J\left( \delta \boldsymbol{\zeta },{^I\mathbf{p}_{1:n}} \right) =\sum_{i=1}^n{J_i\left( \delta \boldsymbol{\zeta },{^I\mathbf{p}_i} \right)}$. Since each component of the objective function is only related to one LiDAR point, the Hessian matrix of the objective function with respect to the increment of the pose can be decomposed into the following summation form:
\begin{equation}
	\frac{\partial ^2J\left( \delta \boldsymbol{\zeta },{^{I}\mathbf{p}_{1:n}} \right)}{\partial \delta \boldsymbol{\zeta }^2}=\sum_{i=1}^n{\frac{\partial ^2J_i\left( \delta \boldsymbol{\zeta },{^{I}\mathbf{p}_i} \right)}{\partial \delta \boldsymbol{\zeta }^2}}
	\label{eq:d2_j_d2_xi_sum}
\end{equation}
Denote $\varepsilon_i=\mathbf{u}_{i}^{\mathrm{T}}(
{^{W}_{I}\tilde{\mathbf{R}}^{*}}\mathrm{Exp}(\delta\boldsymbol\phi){^I\mathbf{p}_{i}}
+{^{W}_{I}\tilde{\mathbf{t}}^{*}}+\delta\mathbf{t}-\mathbf{q}_{i})$, then based on the chain rule and the Jacobian matrix of $\mathrm{SO}(3)$ element, the gradient of the objective function with respect to the increment is given by
\begin{equation}
	\begin{aligned}
		&\frac{\partial J_i\left( \delta \boldsymbol{\zeta },{^{I}\mathbf{p}_i} \right)}{\partial \delta \boldsymbol{\zeta }}=\left( \frac{\partial \varepsilon _i}{\partial \delta \boldsymbol{\zeta }} \right) ^{\mathrm{T}}\frac{\partial J_i\left( \delta \boldsymbol{\zeta },{^{I}\mathbf{p}_i} \right)}{\partial \varepsilon _i}=2{\mathbf{J}_{\varepsilon,i}^{\mathrm{T}}}\varepsilon _i
	\end{aligned}
	\label{eq:d_j_d_xi}
\end{equation}
\begin{equation}
	\mathbf J_{\varepsilon,i}
	=\frac{\partial\varepsilon_i}{\partial\delta\boldsymbol\zeta}
	=\left[\mathbf u_i^{\mathrm T},-\mathbf B_i{^I\mathbf p_i^{\land}}\right], \quad \mathbf B_i=\mathbf u_i^{\mathrm T}{^{W}_{I}\tilde{\mathbf R}^{*}}
\end{equation}
Then the Hessian matrix is
\begin{equation}
	\frac{\partial ^2J_i\left( \delta \boldsymbol{\zeta },^I\mathbf{p}_i \right)}{\partial \delta \boldsymbol{\zeta }^2}=2\mathbf{J}_{\varepsilon,i}^{\mathrm{T}}\mathbf{J}_{\varepsilon,i}
	\label{eq:d2_j_d2_xi}
\end{equation}
Based on \eqref{eq:d_j_d_xi}, it yields
\begin{equation}
	\frac{\partial ^2J_i\left( \delta \boldsymbol{\zeta },^I\mathbf{p}_i \right)}{\partial ^I\mathbf{p}_i\partial \delta \boldsymbol{\zeta }}=2\left[ \begin{array}{c}
		\mathbf{u}_i\mathbf{B}_i\mathrm{Exp}\left( \delta \boldsymbol{\phi } \right)\\
		^I\mathbf{p}_{i}^{\land}\mathbf{B}_{i}^{\mathrm{T}}\mathbf{B}_i\mathrm{Exp}\left( \delta \boldsymbol{\phi } \right) -\left( \mathbf{B}_{i}^{\mathrm{T}}\varepsilon _i \right) ^{\land}\\
	\end{array} \right] 
	\label{eq:d_j_d_p_d_xi}
\end{equation}
Finally, at ${^I\mathbf p_i}={^I\tilde{\mathbf p}_i}$ and $\delta\boldsymbol\zeta=\mathbf0$
, combining \eqref{eq:d2_j_d2_xi_sum}, \eqref{eq:d2_j_d2_xi}, and \eqref{eq:d_j_d_p_d_xi}, the final Jacobian matrix \eqref{eq:d_f_d_p_original} can be given.

\bibliographystyle{IEEEtran}
\bibliography{IEEEabrv,myref}

\end{document}